%% file: main.tex
\newif\ifSingleColumn
\newif\ifShowManuRevisions
\newif\ifShowOverviewResponse
\ShowManuRevisionsfalse
\ShowOverviewResponsetrue
\SingleColumnfalse
\ifSingleColumn
\documentclass[draftclsnofoot, onecolumn, 12pt]{IEEEtran}
\else
\documentclass[twocolumn,journal,10pt]{IEEEtran}
\fi

\usepackage{hyperref}

\input{header.tex}

\input{defines.tex}

\begin{document}

\ifSingleColumn
\title{\linespread{1.2}\huge{Distillation-Enabled Knowledge Alignment for Generative Semantic Communications\\ of AIGC Images}}
\else
\title{\linespread{1.2}\huge{Distillation-Enabled Knowledge Alignment for Generative Semantic Communications of AIGC Images}}
\fi

\author{\linespread{1.25}
\IEEEauthorblockN{
\normalsize{Jingzhi~Hu},~\IEEEmembership{\normalsize Member,~IEEE} and
\normalsize{Geoffrey Ye Li},~\IEEEmembership{\normalsize Fellow,~IEEE} 
}
\thanks{This work has been submitted to the IEEE for possible publication. Copyright may be transferred without notice, after which this version may no longer be accessible.}
\thanks{J. Hu and G. Y. Li are with the Department of Electrical and Electronic Engineering, Imperial College London, London SW7 2AZ, UK.~(email: jingzhi.hu518@gmail.com, geoffrey.li@imperial.ac.uk)}
}

\maketitle
\begin{abstract}
\input{./sections/1_abstract.tex}

\end{abstract}

\ifSingleColumn
\newpage
\fi

\input{sections/2_introduction.tex}

\input{sections/3_system_model.tex}

\input{sections/4_prob_formulate.tex}

\input{sections/5_alg_design.tex}

\input{sections/5_2_implement.tex}

\input{sections/6_simulation.tex}

\input{sections/7_conclusion.tex}

\balance
\renewcommand{\refname}{References} 

\bibliographystyle{IEEEtran}
\bibliography{bibilio}

\end{document}

%% file: header.tex
\usepackage[utf8]{inputenc}
\usepackage{glossaries}
\usepackage[symbols,nogroupskip]{glossaries-extra}
\usepackage{bm}
\usepackage{amssymb}
\usepackage{amsmath}
\usepackage{amsfonts}
\usepackage{mathrsfs}
\usepackage{bbm}
\usepackage{scalerel} 
\usepackage{graphicx}
\usepackage{cite}
\usepackage{enumerate}
\usepackage{url}
\usepackage{epstopdf}
\usepackage{algorithm}
\usepackage[noend]{algpseudocode}
\usepackage{algorithmicx,algorithm}
\usepackage{float}
\usepackage{amsthm}
\usepackage{color}
\usepackage{enumitem}
\usepackage{clipboard}
\usepackage[normalem]{ulem} 
\usepackage{framed}
\usepackage{array} 
\usepackage{hhline}
\usepackage[table]{xcolor}
\usepackage{balance}
\usepackage[caption=false,font=footnotesize]{subfig}
\usepackage{tikz}
\usetikzlibrary{calc}
\usetikzlibrary{decorations.pathreplacing,calligraphy}
\usetikzlibrary{tikzmark}

\usepackage{xargs} 
\usepackage{cooltooltips}
\usepackage{xcolor}  
\usepackage{comment}
\usepackage{xspace}

\usepackage[most]{tcolorbox}

\newtcolorbox{reviewcomment}[1][]{
  colback=blue!5,
  colframe=blue!70!black,
  coltitle=white,
  fonttitle=\bfseries,
  title=#1,
  arc=3pt,
  boxrule=0.8pt,
  left=6pt,
  right=6pt,
  top=6pt,
  bottom=6pt,
}

\newtcolorbox{subreviewcomment}[1][]{
  enhanced,
  breakable,
  frame hidden,   
  title=#1,
  colbacktitle=black!75,    
  coltitle=white,           
  fonttitle=\bfseries,
  boxed title style={
    boxrule=0pt,
    sharp corners,
    left=6pt, right=6pt, top=3pt, bottom=3pt,
  },
  colback=black!5,            
  boxrule=0pt,
  sharp corners,
  left=8pt, right=8pt, top=6pt, bottom=6pt,
}

\newcommand{\beq}{\begin{equation}}
\newcommand{\eeq}{\end{equation}}
\newcommand{\beqq}{\begin{equation*}}
\newcommand{\eeqq}{\end{equation*}}
\newcommand{\beaq}{\begin{align}}
\newcommand{\eeaq}{\end{align}}	
\newcommand{\ssim}{\!\sim\!}
\newcommand{\figwidtha}{0.33\linewidth}
\newcommand{\figwidthb}{0.32\linewidth}
\newcommand{\figwidthc}{0.32\linewidth}

\allowdisplaybreaks

\newcommand{\rev}{\textcolor[rgb]{0,0,0}}

\newenvironment{revenv}
{\begingroup\color[rgb]{0,0,0}}
{\endgroup}

\newcommand{\manuhspacing}{0.08\linewidth}

%% file: defines.tex
\newcommand{\iu}{\mathrm{i}}

\newcommand{\name}{DeKA-g\xspace}
\newcommand{\makd}{MAKD\xspace}

\newcommand{\cala}{CALA\xspace}

\newcommand{\bms}{\bm s}

\newcommand{\bmn}{\bm n}

\newcommand{\bmB}{\bm B}
\newcommand{\bmA}{\bm A}

\newcommand{\expect}{\mathop{\mathbb E}}

\newcommand{\bmf}{\bm f}
\newcommand{\bmg}{\bm g}
\newcommand{\bmG}{\bm G}

\newcommand{\bmu}{\bm u}
\newcommand{\bmv}{\bm v}

\newcommand{\bmz}{\bm z}

\newcommand{\bmx}{\bm x}

\newcommand{\bmM}{\bm M}

\newcommand{\bbR}{\mathbb R}

\newcommand{\cX}{\mathcal X}

\newcommand{\cZ}{\mathcal Z}
\newcommand{\cU}{\mathcal U}

\newcommand{\cD}{\mathcal D}

\newcommand{\cS}{\mathcal S}

\newcommand{\cG}{\mathcal G}

\newcommand{\cN}{\mathcal N}

\newcommand{\rG}{\mathrm G}

\newcommand{\rJ}{\mathrm J}

\newcommand{\reu}{\mathrm{EU}} 
 
\newcommand{\rLE}{\mathrm{LE}} 
\newcommand{\rLD}{\mathrm{LD}} 
 
\newcommand{\rJE}{\mathrm{JE}} 
\newcommand{\rJD}{\mathrm{JD}} 
\newcommand{\rGSC}{\mathrm{GSC}}

\newcommand{\rCG}{\mathrm{CG}} 
\newcommand{\rEG}{\mathrm{EG}}

\newcommand{\rjscc}{\mathrm{jscc}}
\newcommand{\rgen}{\mathrm{gen}}
\newcommand{\gen}{\bm f_{\mathrm{G}}}

\newcommand{\figwidth}{0.33\linewidth}

%% file: sections/1_abstract.tex
Due to the surging amount of AI-generated images, its provisioning to edges and mobile users from the cloud incurs substantial traffic on networks. Generative semantic communication (GSC) offers a promising solution by transmitting highly compact information, i.e., prompt text and latent representations, instead of high-dimensional image data. However, GSC relies on the alignment between the knowledge in the cloud generative AI~(GAI) and that possessed by the edges and users, and between the knowledge for wireless transmission and that of actual channels, which remains challenging. In this paper, we propose DeKA-g, a distillation-enabled knowledge alignment algorithm for GSC systems. The core idea is to distill the image generation knowledge from the cloud-GAI into low-rank matrices, which can be incorporated by the edge and used to adapt the transmission knowledge to diverse wireless channel conditions. DeKA-g comprises two novel methods: metaword-aided knowledge distillation (MAKD) and \rev{condition-aware low-rank adaptation (CALA)}. For MAKD, an optimized metaword is employed to enhance the efficiency of knowledge distillation, while \rev{CALA enables efficient adaptation to diverse rate requirements and channel conditions}. From simulation results, DeKA-g improves the consistency between the edge-generated images and the cloud-generated ones by $44$\% and \rev{enahnces the average transmission quality in terms of PSNR by 6.5~\!dB over the baselines without knowledge alignment}.

%% file: sections/2_introduction.tex
\section[Introduction]{Introduction}
\label{SEC_INTRO}

The rapid advancement of artificial intelligence (AI) has led to the phenomenal success of generative AI (GAI), exemplified by GPT~\cite{Bubeck2023Arxiv_Sparks}, Stable Diffusion~\cite{Rombach22CVPR_High}, and SoRA~\cite{sora2024}, which fundamentally changes how people work and live~\cite{Mckinsey_report,Erik25QJE_GAI}.
The essential capability of GAI models is to generate high-quality synthetic content, which is derived from learning the underlying patterns and distributions of massive examples~\cite{Jovanovic22GAI}.
With their generation capabilities, GAI models are able to produce content of various modalities in both high quality and large quantity.
Moreover, with the ubiquitous connectivity of the Internet and wireless networks, GAI models deployed on cloud servers can provision a vast amount of high-quality AI-generated content (AIGC) data to users worldwide, unlocking an unprecedented level of abundance of resources for entertainment and productivity.

However, provisioning a large volume of high-dimensional AIGC data for edge servers and users also places a significant burden on both the Internet and wireless networks. 
In particular, given the sheer volume of AI-generated images, AIGC image traffic can severely compete with conventional traffic, resulting in network congestion and degraded user experience.
Fortunately, the emerging semantic communication~(SC) paradigm offers a promising solution to reduce the traffic of AIGC image provisioning. 
The principle of SC is to leverage deep neural models to encode high-dimensional data into low-dimensional representations that preserve only the essential semantic information and transmit them to the receiver~\cite{Xie21TSP_Deep}.
In stark contrast to conventional data compression techniques, SC does not require precise reconstruction of data at the receiver but focuses on the effectiveness of transmission to the target task, thereby allowing more aggressive size reduction and improved transmission efficiency.

Notably, SC is naturally suitable for AIGC image provisioning{,} as AIGC images are derived from low-dimensional semantic information.
{Several studies have explored} the transmission of AIGC images using SC.
In~\cite{Liu24TMC_Cross}, cross-modal attention maps extract the semantic information from AIGC images based on their corresponding prompt texts.
{In} the cloud-edge-user framework in~\cite{Xia25WCM_GAI}, AIGC images are generated in the cloud, sent to wireless users by the edge server via SC, and refined by the users' local GAI models. 
In addition, a few studies integrate GAI models into SC to enhance data reconstruction, leading to the emergence of generative SC~(GSC) systems.
The generative joint source and channel coding~(JSCC) scheme in~\cite{Erdemir23JSAC_Generative} incorporates a StyleGAN-2 GAI model into the JSCC decoder to reconstruct high-quality images from semantic features.
Following the same spirit, efficient frameworks are developed for integrating various types of GAI models into JSCC decoders~\cite{Liu24Netw_Semantic,Wang24Netw_Harness,Cheng25TMC_Wireless}.
The SC framework in~\cite{Grassucci24Av_Rethink} tailors diffusion models for multi-user scenarios. 
In~\cite{Lin23Netw_Unified}, an integrated SC and AIGC framework is designed for metaverse applications, focusing on network resource allocation and user coordination.
In~\cite{Liang25TCCN_GAI}, recent works on the GAI-integrated SC are comprehensively summarized, focusing on communication schemes and aspects of network management. 

\ifSingleColumn
    \renewcommand{\figwidth}{0.7\linewidth}
\else
    \renewcommand{\figwidth}{1\linewidth}
\fi
\begin{figure}[t] 
    \centering
    \includegraphics[width=\figwidth]{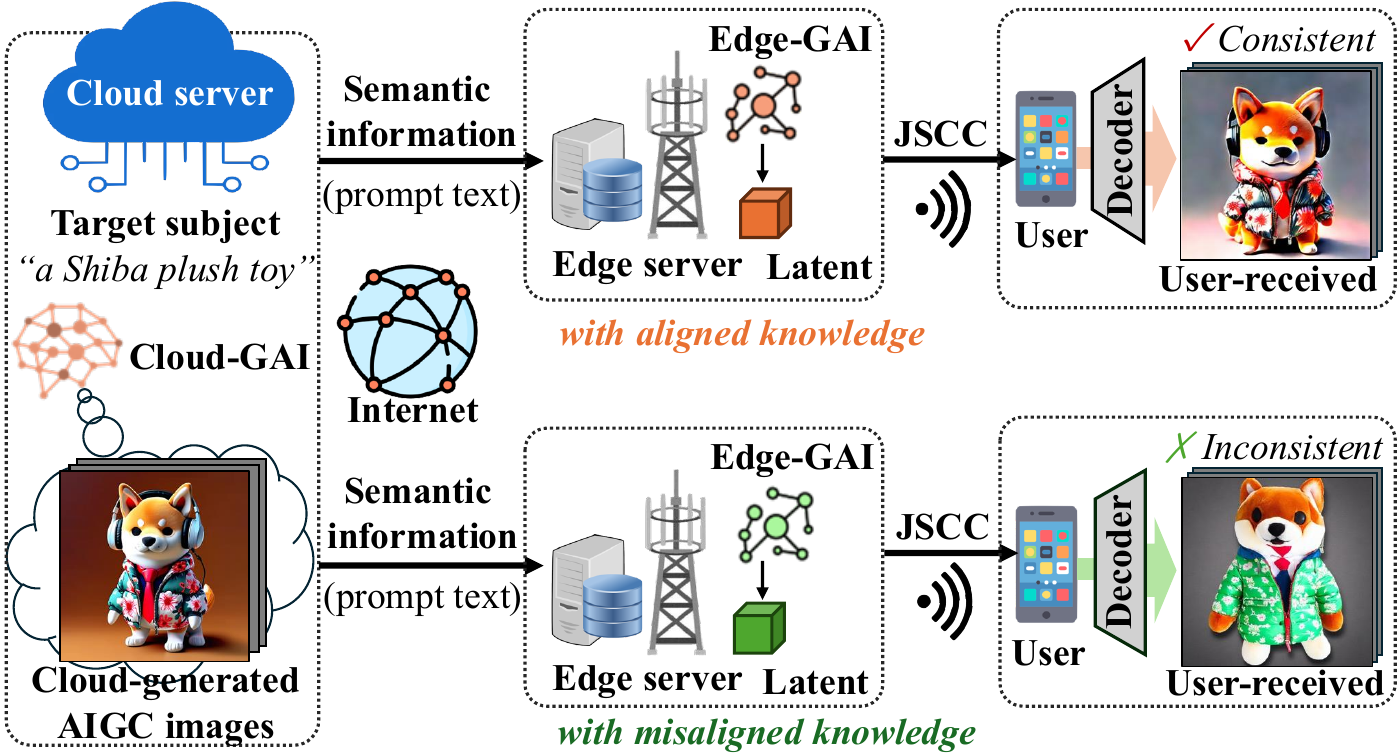}
    \vspace{-2ex}
    \caption{The GSC system for AIGC image provisioning, under aligned and misaligned knowledge.}
    \label{fig_illu_1}
\end{figure}

\Copy{R2-1-1}{\rev{In this work, we focus on the AIGC provisioning task under a three-tier cloud-edge-user architecture.
Each tier in the cloud-edge-user architecture plays a vital role in {the} AIGC provisioning task over {the} network. 
The cloud hosts large-scale GAI models with rich generation knowledge to meet user demands, while the user is the consumer of AIGC content.
The edge provides distributed computing resources in close proximity to users and has been widely recognized as a key enabler of efficient networks~\cite{EdgeIntelligence_6GWhitePaper,3GPP_TR23748_MEC}.
Notably, modeling the full cloud-edge-user architecture is important.
A simplified cloud-edge structure overlooks the wireless network between the edge and users, which often constitutes the performance bottleneck~\cite{Zangooei23COMMAG_RAN}. 
In contrast, a cloud-user structure will place all computation in the cloud, leading to excessive network traffic, heavy cloud workload, and high end-to-end latency{,} especially {at} a large user scale~\cite{EdgeIntelligence_6GWhitePaper}.}}

In the three-tier architecture in~\cite{Xia25WCM_GAI}, the edge leverages GSC to boost the wireless transmission between {the edge} and {the user}. 
However, the cloud-to-edge transmission in~\cite{Xia25WCM_GAI} still follows the conventional {communication} paradigm, imposing a significant burden on the backbone network.
In comparison, we consider a GSC system where the cloud-to-edge transmission also adheres to the SC paradigm.
As shown in Fig.~\ref{fig_illu_1}, instead of transmitting AIGC images, the cloud transmits their semantic information, i.e., the prompt text describing the target subject, to the edge, thereby reducing the traffic load from megabytes to only a few hundred bytes.
The edge leverages its local GAI model, namely the edge-GAI, to generate latent representations of images, and then transmits them to the user via the bandwidth-efficient JSCC scheme.
Finally, the received latent representations are decoded by the user to reconstruct AIGC images.%
\Copy{R2-1_2}{\rev{Built upon this three-tier architecture, our GSC system alleviates the burden on both the backbone and the wireless access 
{network}.}}
\Copy{R1-2}{\rev{Moreover, by exploiting edge-side generation close to the user, it reduces the computation load of the cloud and the service latency 
{for the user}.
{Consequently}, it is suitable for bandwidth-limited backbone networks (e.g., satellite or maritime networks), cloud-to-edge offloading scenarios, and latency-critical AIGC applications.}}

\Copy{R3-4_1}{\rev{For GSC systems, their high efficiency fundamentally depends on the knowledge alignment (KA) among the cloud, edge, and user with the target task~\cite{Liang25TCCN_GAI,Xia25WCM_GAI}.
In particular, the generation knowledge embedded in the edge-GAI should be aligned with that of the cloud-GAI to generate consistent}}\footnote{\Copy{R1-3_1}{\rev{As heterogeneous GAIs inherently produce different images for the same prompt, the consistency here needs to be defined beyond pixel level, which is elaborated in Sec.~\ref{s2ec_prob_form}.}}}
\Copy{R3-4_1_2}{\rev{images, and the transmission knowledge in JSCC should be aligned with the actual transmission conditions and data distribution.
Failures in either will degrade the quality of the user-received images and their consistency with the target ones, as illustrated in Fig.~\ref{fig_illu_1}.
The existing works commonly adopt an implicit assumption that the components of GSC systems are owned by a single entity and constitute a unified system~\cite{Xia25WCM_GAI, Erdemir23JSAC_Generative, Liu24Netw_Semantic,Wang24Netw_Harness,Cheng25TMC_Wireless, Grassucci24Av_Rethink, Lin23Netw_Unified}.
As a result, the GAI models and JSCC modules can be jointly trained for the task, and thus the KA is ensured by default.
However, in practice, neural models across the network are often heterogeneous rather than aligned toward a common objective.
Consequently, the knowledge embedded in the component models of GSC systems can be misaligned with the AIGC image provisioning task, making it imperative to address KA.}}

Existing studies only consider KA in SC systems 
{without} GAI. 
In~\cite{Zhang23JSAC_Deep,Choi24TVT_Semantics,guo2024survey}, fine-tuning is used to align the knowledge embedded in the neural 
{modules} of JSCC.
In~\cite{Sana23GC_Semantic,Fiorellino24Arxiv_Dynamic}, an equalization approach is proposed for aligning different semantic latent spaces. 
Additionally, in our previous work~\cite{Hu25AV_Distillation}, a KA protocol is designed based on knowledge distillation~(KD).
To the best of the authors' knowledge, the KA problem for GSC systems has not been addressed. 
Addressing it is challenging due to the high complexity of GAI models and the difficulty of measuring KA performance given the {inherent} stochasticity of generation processes and wireless channels.

\Copy{R3-4_2}{\rev{
    In this paper, we take a first step toward a tractable optimization formulation of the KA problem in GSC, beyond prior qualitative analyses~\cite{Liang25TCCN_GAI,Xia25WCM_GAI,guo2024survey}.}}
\Copy{R2-0-1}{\rev{The KA problem aligns the prior knowledge across different entities in a GSC system with the task requirement by adapting a compact set of parameters.
It inherently includes two aspects: i) aligning the generation knowledge at the edge with what is needed to produce the target images, and ii) aligning the transmission knowledge at the edge-user link with what is needed to efficiently deliver target images under actual transmission conditions.}}
To solve the KA problem, we propose \emph{\name}, a distillation-enabled KA algorithm.
Stemming from our protocol in~\cite{Hu25AV_Distillation}, we let the cloud distill aligned knowledge into low-rank adaptation~(LoRA) matrices and send them to the edge and the user to achieve KA.
Specifically, two novel methods are introduced in \name for efficient KA of generation and transmission knowledge.
The first method, metaword-aided KD~(\makd), optimizes a metaword to align the prompt text interpretation of the edge-GAI with the cloud-GAI, thus facilitating more effective KD into compact LoRA matrices.
\rev{The second method, condition-aware LoRA~(\cala), endows distilled knowledge with condition adaptiveness by softly reweighting its rank-1 components according to transmission conditions. 
The main contributions of this paper can be summarized below.}
\begin{itemize}[leftmargin=*]
\item \rev{We formulate a tractable KA problem to address the challenging misalignment between the knowledge embedded in a GSC system and that required for consistent generation and efficient transmission of target images.}
\item \rev{We propose the \name algorithm to solve the KA problem, featuring two novel methods: \makd and \cala. The former aligns generation knowledge between the edge-GAI and the cloud-GAI, and the latter enables efficient transmission KA under diverse conditions.}
\item \rev{Based on our simulation results, \name improves the consistency between edge-generated and cloud-generated AIGC images by $44\%$, and enhances the average transmission quality in terms of PSNR by 6.5 dB.}
\end{itemize}

The rest of this paper is organized as follows.
In Sec.~\ref{sec_system_lodel}, we model the GSC system.
In Sec.~\ref{sec_prob_form}, we formulate the KA problem for the GSC system.
In Sec.~\ref{sec_alg_design}, we propose the \name algorithm to solve the KA problem.
We design the neural model implementation of the GSC system in Sec.~\ref{sec_implement} and provide the evaluation results in Sec.~\ref{sec_eval}.
A conclusion is drawn in Sec.~\ref{sec_conclu}.

%% file: sections/3_system_model.tex
\section[System Model]{GSC Systems}
\label{sec_system_lodel}

We model the GSC system for AIGC image provisioning, which comprises a cloud server, an edge server, and a mobile user.
The mobile user connects to the wireless network hosted by the edge server, while the edge server connects to the cloud server via the Internet.
Both the cloud and the edge servers are equipped with GAI models, namely the cloud-GAI and {the} edge-GAI.
The goal of the AIGC image provisioning task is to provide the user with {a substantial number of} AIGC images of a target subject, while ensuring high consistency between the user-received images and those generated by the {cloud-GAI}.

For efficiency, the operation of the GSC system should follow two key principles:
(i) the traffic load on both the Internet and the wireless network should be minimized; (ii) the computational and storage burden on the user should be low, as the user has {fewer resources} than the edge server and the {cloud server}.
\Copy{R2-4}{\rev{Accordingly, we model the procedures for the GSC system to perform the AIGC image provisioning task as follows, which is illustrated as Fig.~\ref{fig_syst_diag}.
First, the cloud server sends the semantic information of the target subject, i.e., its detailed prompt text, to the edge server.
With the prompt, the edge-GAI, which we refer to as the \emph{generation module}, converts the prompt into latent {representations} (\emph{latents} for short).
The generated latents are then forwarded to the JSCC codec, referred to as the \emph{transmission module}.
The latents are mapped to symbols, transmitted to the user over the noisy wireless channel, and mapped back to an image in the pixel space.}}
We then elaborate on the generation module and the transmission module below.

\ifSingleColumn
    \renewcommand{\figwidth}{0.7\linewidth}
\else
    \renewcommand{\figwidth}{1\linewidth}
\fi
\begin{figure}[t] 
\Copy{R2-4_1}{
    \centering
    \includegraphics[width=\figwidth]{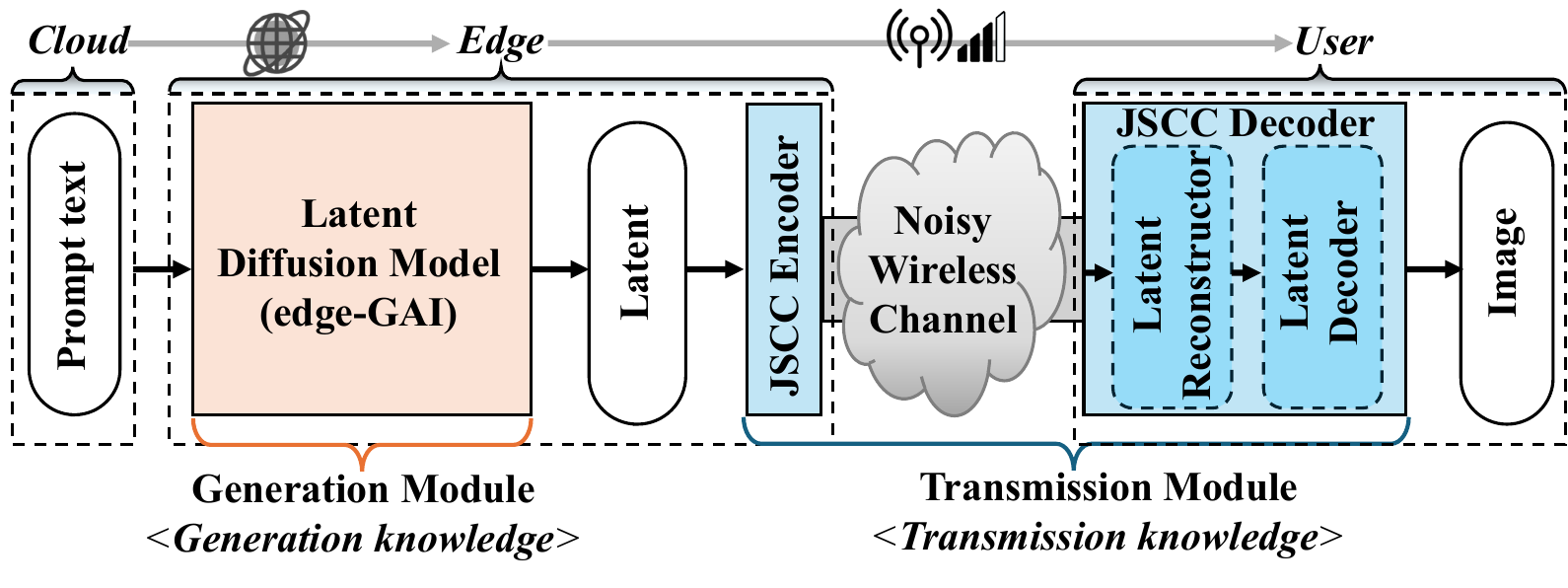}
    \vspace{-3ex}
    \caption{\rev{Systematic diagram of the GSC system.}}
    \label{fig_syst_diag}}
\end{figure}

\input{sections/3_a_GAI_model.tex}

\input{sections/3_b_SC_trans.tex}

%% file: sections/3_a_GAI_model.tex
\subsection{Generation Module}\label{s2ec_aigc_model}

We assume that the edge-GAI is a latent diffusion model~(LDM)~\cite{Rombach22CVPR_High} due to its efficiency in generating high-quality images and proven effectiveness in GSC systems~\cite{Lin23Netw_Unified,Wang24Netw_Harness,Cheng25TMC_Wireless}.
We denote the edge-GAI by $\gen:(\cU,\cZ)\rightarrow\cZ$, which takes a prompt text in text space $\cU$ and a random noise in latent space $\cZ$ as input and outputs a latent in $\cZ$.
Following~\cite{Rombach22CVPR_High}, latent space $\cZ$ is established by a latent encoder--decoder pair, denoted by $\bm f_{\rLE}(\cdot)$ and $\bm f_{\rLD}(\cdot)$, which map pixel images to latents and vice versa.
For more compact and efficient representations, the dimensionality of the latent space is much smaller than that of the pixel space.
As a result, the generation in the latent space is more efficient than that in the pixel space.

In principle, the edge-GAI assumes that the input noise is a noisy version of an underlying noiseless latent after $T$ steps of diffusion processes.
Then, leveraging the prompt text as conditional information, it performs an iterative denoising process to recover the noiseless latent.
The core module of the edge-GAI is a \emph{noise predictor}, denoted by $\bm{\varepsilon} : (\cZ, \mathbb Z, \cU) \rightarrow \cZ$, where $\mathbb{Z}$ denotes the integer space of diffusion steps.
The noise predictor takes the noisy latent $\bm{z}_t\in\cZ$, its diffusion step $t \in \mathbb{Z}$, and prompt text $\bm{u} \in \mathcal{U}$ as inputs and outputs the predicted noise component $\bm{\epsilon}\in\cZ$ in $\bm{z}_t$.

\begin{revenv}
The capability of the noise predictor stems from both its neural model architecture and its parameter optimization.
In general, the training process is conducted on a comprehensive dataset $\cD_{\rgen}$ comprising latent-text pairs of a large collection of images.
The parameters of the noise predictor are optimized for the minimization of the mean-squared error (MSE) between the predicted noise and the actual noise at an arbitrary diffusion step $t$.
Without loss of generality, we denote the parameters of the noise predictor as $\varTheta_{\rG}$, represented as a collection of real-valued matrices. 
Then, the training can be expressed as the following optimization problem:
\ifSingleColumn
\begin{align}\label{equ_opt_noise_est}
\min_{\varTheta_{\rG}}~
\sum_{(\bmz,\bmu)\in\cD_{\rgen}}\mathop{\mathbb E}\limits_{t,\bm\epsilon}\Big[\|\bm\epsilon - \bm\varepsilon(\bm z_t, t, \bmu; \varTheta_{\rG})\|^2\Big],
\end{align}
\else
\begin{align}\label{equ_opt_noise_est}
\min_{\varTheta_{\rG}}~
\sum_{(\bmz,\bmu)\in\cD}\mathop{\mathbb E}\limits_{t,\bm\epsilon}\Big[\|\bm\epsilon - \bm\varepsilon(\bm z_t, t, \bmu; \varTheta_{\rG})\|^2\Big],
\end{align}
\fi
where $\bm\epsilon$ is randomly sampled from a normal distribution, $\bmz_t=\sqrt{\alpha_t}\bm z + \sqrt{1-\alpha_t}\cdot \bm\epsilon$, and $\alpha_t$ is a diffusion coefficient specified by the noise schedule in~\cite{DDIM}.
\end{revenv}

%% file: sections/3_b_SC_trans.tex
\subsection{Transmission Module}\label{s2ec_tx_mod}

The edge server transmits the generated latents to the user via JSCC, with the JSCC encoder and decoder deployed at the edge and user sides, respectively.
The JSCC encoder $\bmf_{\rJE}:\cZ\rightarrow \mathbb{C}^K$ takes a latent ${\bmz} \in \cZ$ as input and outputs a complex symbol vector $\bms \in \mathbb{C}^K$, where $K$ is the symbol length.
The JSCC decoder $\bmf_{\rJD}:\mathbb{C}^K\rightarrow\cX$ takes received symbol vectors as input and outputs images. 
Specifically, let $X$ denote the number of pixels in an image. 
We define the \emph{rate} as the number of symbols per pixel, i.e., $\tau=K/X$, where a higher rate implies more spectral resources to be used. 
In practice, the rate should be determined by resource scheduling and allocation of the wireless network, and is therefore considered as a transmission condition.

Directly transmitting $\bms$ is incompatible with practical hardware, which only supports a discrete set of complex symbols known as constellation points.
\rev{To address this issue, we quantize each symbol in $\bms$ to its nearest constellation point as in~\cite{DeepJSCC_Q}. 
The quantized symbol vector is denoted by $\hat{\bms}$ and transmitted via orthogonal frequency division multiplexing (OFDM).}
In OFDM, the channel is divided into $J$ sub-channels, each carrying one symbol in a time slot.
Without much loss of generality, we assume that symbols of $\hat{\bms}$ are allocated to the sub-channels sequentially.
Then, based on~\cite{Jiang22JSAC_Wireless,DeepJSCC_Q}, the received signal for the $k$-th symbol of $\hat{\bms}$ is
\beq\label{equ_ch}
y_k = h_j\cdot \hat{s}_k + \delta, 
\eeq
where $h_j$ denotes the Rayleigh channel gain of the $j$-th sub-channel, $\delta$ denotes the channel noise with power $P_{\delta}$.
The symbols allocated to the same sub-channel are transmitted in sequential time slots, where the gain of each sub-channel is constant as in the block-fading channel model.

In contrast to the uncorrelated sub-channel gains as in~\cite{Jiang22JSAC_Wireless,DeepJSCC_Q}, we consider correlated sub-channel gains, which are more realistic.
Based on~\cite{goldsmith2005wireless} and~\cite{Barriac06TCOM_Space}, the correlation between $h_{j'}$ and $h_{j}$~($\forall j',j\in\{1,...,J\}$) can be modeled as
\beq\label{equ_cov_mat}
C_{j'j} = \mathbb E[h_{j'}{h_{j}^*}] = P_{\mathrm{h}}/ \exp(1+ 2\pi\iu\cdot\Delta_{j'j}\cdot\omega), 
\eeq
where $h_{j}^*$ denotes the conjugate of $h_{j}$, $P_{\mathrm{h}}$ is the average power, $\omega$ denotes the \emph{delay spread} of the power delay profile, and $\Delta_{j'j}$ is the frequency spacing between the $j'$-th ad the $j$-th sub-channels.
Consequently, the gain vector for the sub-channels, i.e., $\bm h = (h_1, \dots, h_J)$, follows a complex Gaussian distribution $\mathcal{CN}(\bm{0}, \bm{C})$, where the elements of $\bm{C}$ are given by~\eqref{equ_cov_mat}.
For simplicity of notation, we refer to $\gamma=P_{\mathrm{h}}/P_{\delta}$ as the \emph{signal-to-noise ratio~(SNR)} of the wireless channel.

Upon receiving $y_1,\dots, y_K$, the user estimates the sub-channel gains and applies equalization to recover the transmitted symbols.
\rev{The recovered symbols are denoted by $\check{\bms}=(\check{s}_1,...,\check{s}_K)$ and sent to the JSCC decoder.}
In particular, we assume that the JSCC decoder consists of two sub-modules, i.e., a latent reconstructor and a latent decode. 
The latent reconstructor maps $\check{\bms}$ back to latent $\check{\bmz}$ in the latent space, and the latent decoder maps $\check{\bmz}$ to image $\check{\bmx}$.

\begin{revenv}
To handle various latents under diverse transmission conditions, the parameters of the JSCC codec need to be trained on a comprehensive dataset.
Such a dataset, denoted by $\cD_{\rjscc}$, should comprise a substantial number of image-latent pairs.
Moreover, we focus on three types of transmission conditions, including rate $\tau$, SNR $\gamma$, and delay spread $\omega$, which together capture key aspects of resource scheduling and channel characteristics.
We represent a transmission condition by $\bm\phi=(\tau,\gamma,\omega)$ and denote the distribution of $\bm\phi$ by $\varPhi$.
Based on above notations, the training can be expressed as
\beq\label{equ_pt_tx}
\min_{\varTheta_{\rJ}}~\underset{\bm\phi \sim \varPhi}{\mathbb{E}}\sum_{(\bmx,\bmz)\in \cD_{\rjscc}} L(\bmx,\bmz; \varTheta_{\rJ}|\bm\phi),
\eeq
where $\varTheta_{\rJ}$ is the parameters of the JSCC codec, and
$L(\bmx,\bmz; \varTheta_{\rJ}|\bm\phi)$ denotes the objective function.
We adopt a representative objective function composed of four components: (i) the MSE between the reconstructed and the original latent; (ii) the MSE between the received and the original image;
(iii) the perceptual loss measured by the distance between extracted features from the received and original images by a pre-trained visual neural model; and
(iv) a regularization loss measured by the distance between the encoded symbols and their nearest constellation points.
\end{revenv}

\vspace{1ex}
In summary, the complete AIGC image provisioning process in the GSC system can be represented as a composite function $\bmf_{\rGSC}: (\cU,\cZ) \rightarrow \mathcal{X}$, which maps the prompt text of the target subject and a random noise input to a received image.
By varying the noise input, the GSC system can provide the user with numerous AIGC images of the same target subject.

%% file: sections/4_prob_formulate.tex
\section{KA Problem Formulation for GSC Systems}\label{sec_prob_form}
In this section, we first explain the motivation behind the KA in GSC systems.
Then, we formulate the KA problem as an optimization problem and highlight its key challenges.
\subsection{Motivation}\label{ssec_prob_motiv}
According to Sec.~\ref{sec_system_lodel}, the core components of the GSC system are the noise predictor $\bm{\varepsilon}(\cdot)$ and JSCC codec $\bmf_{\rJE}(\cdot)$ and $\bmf_{\rJD}(\cdot)$, all of which are deep neural models with large numbers of parameters.
The functionality of the GSC system depends on two types of knowledge embedded in the parameters: the generation knowledge learned by $\bm{\varepsilon}(\cdot)$ through the training in \eqref{equ_opt_noise_est}, and the transmission knowledge embedded in $\bmf_{\rJE}(\cdot)$ and $\bmf_{\rJD}(\cdot)$ via the training in \eqref{equ_pt_tx}.

\Copy{R2-3_2}{\rev{However, in practice, the edge-GAI's generation knowledge is often misaligned with what is needed to generate images consistent with the cloud-GAI.
This misalignment can be attributed to differences in the neural model scale, architecture, training strategies, and training dataset between the cloud and edge GAIs.
Even if the edge-GAI is an ideally distilled or compressed cloud-GAI, the continuous updates of the cloud-GAI over time can reintroduce misalignment of generation knowledge. 
As a result, generation knowledge misalignment is pervasive and persistent, recognized as a key challenge in SC~\cite{guo2024survey, Xia25WCM_GAI, Yang22WCM_Semantic,Liang24TCCN_Generative}.}}
An analog scenario in human communication can help illustrate this misalignment: when one person describes a subject to another, even with precise wording, the listener may envision a distinctly different subject.

\rev{Moreover, the transmission knowledge in the JSCC codec often misaligns with both the actual data distribution and the transmission conditions.
On the one hand, the JSCC codec is typically trained to optimize the average performance over diverse latents, rather than being tailored to the latents of target subject.
On the other hand, the distribution of transmission conditions assumed in training may differ from the actual ones of the edge-to-user transmission.
As a result, the JSCC codec becomes inefficient, degrading the quality of the user-received images and further exacerbating their inconsistency with the cloud-generated ones.}
Therefore, achieving KA for both generation and transmission knowledge is important for the GSC system.

\subsection{Problem Formulation}\label{s2ec_prob_form}
Despite its importance, formulating the KA as a tractable optimization problem remains non-trivial.
The first challenge is how to define and quantify the consistency between images received by users and images generated by the cloud-GAI.
Although both are generated from the same prompt text of the same target subject, significant structural and visual discrepancies generally exist in the images generated by different GAI models.

\Copy{R1-3_2}{\rev{To address this difficulty, we define the consistency by the sum of \emph{visual similarity} and \emph{semantic similarity} as in~\cite{Ruiz23CVPR_Dreambooth}. 
Visual similarity is computed as the cosine similarity between features extracted by the DINO model~\cite{Caron21ICCV_Emerging}, as DINO provides multi-scale visual features that are stable for perceptual comparison.
Semantic similarity is computed using cosine similarity between features extracted by the CLIP model~\cite{Radford21ICML_Learning}, as CLIP embeds images into a shared semantic space that captures high-level meaning.
Accordingly, we also refer to the visual and semantic similarities as the \emph{DINO-score} and \emph{CLIP-score}, respectively.}}

\Copy{R2-2b-1}{\rev{To formulate a tractable KA problem, it is also important to identify a feasible KA procedure considering practical constraints.
Achieving KA generally requires training and updating the parameters of both the generation and transmission modules. 
However, directly training these modules at the edge or the user sides can be impractical due to their limited computational resources~\cite{EdgeIntelligence_6GWhitePaper} and the substantial training cost of GAI models.}
\rev{Typically, neural models of the edge and the user are often deployed from the cloud~\cite{Tian24EdgeCloudGenAI,Xu23GLOBECOM_Joint}; therefore, the cloud may retain reference copies of the models for maintenance, supported by its ample storage spaces~\cite{EdgeIntelligence_6GWhitePaper}.
In light of this, we assume that the cloud computes the parameter updates for KA and then sends them to the edge and the user.
Moreover, to minimize network overhead, the parameter updates should be highly compact, i.e., the number of parameters in the updates should be significantly smaller than that of the original modules.}}

\Copy{R2-2b-2}{\rev{Based on the above KA procedure, the core scientific problem is: 
{given a prompt text of a target subject, under a strict size budget for the update parameters, how to maximize the consistency between the images generated by the cloud GAI and the edge-generated images received by the user?}}}
Accordingly, denoting the set of update parameters for the noise predictor by $\Delta\varTheta_{\rG}$ and that for the JSCC codec by $\Delta\varTheta_{\rJ}$, we formulate the KA problem as
\ifSingleColumn 
\rev{\begin{align}\label{equ_ka_prob}
\max_{\Delta \varTheta_{\rG}, \Delta \varTheta_{\rJ}} \quad 
&\expect\limits_{\substack{\bmn,\bmn'\sim\cN\\ \bm\phi \sim \varPhi_{\reu}}}\left[
S_{\mathrm{DINO}}\big(\bmf_{\rGSC}(\bmv,\bmn),\bmf_{\rCG}(\bmv,\bmn')\big)+S_{\mathrm{CLIP}}\big(\bmf_{\rGSC}(\bmv,\bmn),\bmf_{\rCG}(\bmv,\bmn')\big)\right], \\
\text{s.t.}\quad & |\Delta \varTheta_{\rG}|\ll |\varTheta_{\rG}|,~|\Delta \varTheta_{\rJ}| \ll |\varTheta_{\rJ}|, \nonumber
\end{align}}
\else
\rev{\begin{align}\label{equ_ka_prob}
\max_{\Delta \varTheta_{\rG}, \Delta \varTheta_{\rJ}} \quad 
&\expect\limits_{\substack{\bmn,\bmn'\sim\cN\\ \bm\phi \sim \varPhi_{\reu}}}\Big[
S_{\mathrm{DINO}}\big(\bmf_{\rGSC}(\bmv,\bmn),\bmf_{\rCG}(\bmv,\bmn')\big) \\[-5pt]
&\qquad\qquad +S_{\mathrm{CLIP}}\big(\bmf_{\rGSC}(\bmv,\bmn),\bmf_{\rCG}(\bmv,\bmn')\big)\Big], \nonumber\\
\text{s.t.}\quad & |\Delta \varTheta_{\rG}|\ll |\varTheta_{\rG}|,~|\Delta \varTheta_{\rJ}| \ll |\varTheta_{\rJ}|,  \nonumber
\end{align}}
\fi
\noindent where $\bm v$ is the prompt text of the target subject, $\bmn,\bmn'$ are the noise sampled from the normal distribution, $S_{\mathrm{DINO}}(\cdot)$ and $S_{\mathrm{CLIP}}(\cdot)$ are the DINO-score and CLIP-score, respectively, $\bmf_{\rCG}(\bmv,\bmn')$ represents the image generation of the cloud-GAI, $\varPhi_{\reu}$ denotes the actual distribution of transmission conditions, and $|\cdot|$ denotes the set cardinality, i.e., the number of elements.

%% file: sections/5_alg_design.tex
\section{Distillation-Enabled KA Algorithm for GSC systems}\label{sec_alg_design}

We design the \name algorithm to solve the KA problem~\eqref{equ_ka_prob}.
\name first decomposes the problem into two sub-problems: the KA of generation (\emph{G-KA}) and the KA of transmission (\emph{T-KA}).
Then, two novel methods are introduced: the \uline{m}etaword-\uline{a}ided \uline{k}nowledge \uline{d}istillation (\makd) method for the G-KA problem, and the \uline{c}ondition-\uline{a}daptive \uline{l}ow-rank \uline{a}daptation (\cala) method for the T-KA problem.

\subsection{Decomposition of KA Problem}\label{s2ec_decompose}

The key idea of \name is to distill generation and transmission knowledge into $\Delta \varTheta_{\rG}$ and $\Delta \varTheta_{\rJ}$.
To achieve this goal efficiently, it decomposes~\eqref{equ_ka_prob} and optimizes $\Delta \varTheta_{\rG}$ and $\Delta \varTheta_{\rJ}$ in turn.
When optimizing $\Delta \varTheta_{\rG}$, the objective of~\eqref{equ_ka_prob} can be interpreted as enabling the edge-GAI to generate latents of cloud-generated images.
Based on KD~\cite{Hinton14NIPS_Distilling}, this can be achieved by directly using latents of cloud-generated images to train the noise predictor.
By this conversion, we also avoid the high computational cost associated with the DINO- and CLIP-scores.

To ensure a compact size of $\Delta\varTheta_{\rG}$, we adopt the LoRA technique~\cite{Hu2021LoRA}.
\rev{Specifically, we let each update of parameter matrices in $\Delta\varTheta_{\rG}$ to be a LoRA matrix with rank $R$ much lower than its size.
Each LoRA matrix $\bmM \in \bbR^{H\times W}$ in $\Delta\varTheta_{\rG}$ can be expressed as $\bmB\bmA$, where $\bm B\in\bbR^{H\times R}$ and $\bm A\in\bbR^{R\times W}$.
We refer to $R$ as the \emph{rank budget} for the update parameters.
}
Accordingly, we convert the optimization of $\Delta\varTheta_{\rG}$ in~\eqref{equ_ka_prob} into the G-KA problem as follows,
\beq
\label{equ_g_ka}
\min_{\Delta\varTheta_{\rG}}~\sum_{\bmz\in\cS_{\rCG}}\expect\limits_{t,\bm\epsilon}\Big[\|\bm\epsilon - \bm\varepsilon(\bmz_t, \bmv, t;\varTheta_{\rG}+\Delta\varTheta_{\rG} )\|^2\Big],
\eeq
where $\cS_{\rCG}$ denotes the training latent set, and other symbols are as defined in~\eqref{equ_opt_noise_est}.

\Copy{R1-1}{\rev{
The set $\cS_{\rCG}$ can be obtained by collecting $N_{\rCG}$ cloud-generated images and encoding them using the latent encoder $\bmf_{\rLE}(\cdot)$ associated with the edge-GAI's latent space. 
Therefore, the G-KA remains feasible even when the cloud GAI employs a latent space defined by a different latent encoder-decoder pair from that of the edge GAI.}}
Conceptually, \eqref{equ_g_ka} distills the subject-related generation knowledge of the cloud GAI into the LoRA matrices.
Owing to their compact size, which enables efficient cloud-to-edge transmission, these LoRA matrices can be optimized at the cloud, where more abundant computational resources are available.

Provided that the G-KA in~\eqref{equ_g_ka} is handled, \eqref{equ_ka_prob} reduces to the KA for latent transmission and decoding, i.e., the T-KA.
Based on~\eqref{equ_pt_tx}, this T-KA problem can be handled by
\beq
\label{equ_t_ka}
\min_{\Delta\varTheta_{\rJ}}~\underset{\bm\phi \sim \varPhi_{\reu}}{\mathbb{E}}\sum_{(\bmx,\bmz)\in \cS_{\rEG}} L(\bmx,\bmz; \varTheta_{\rJ}+\Delta\varTheta_{\rJ}|\bm\phi),
\eeq
where $\Delta\varTheta_{\rJ}$ consists of LoRA matrices for the update of $\varTheta_{\rJ}$,
$\cS_{\rEG}$ is a set of edge-generated latents of the target subject after G-KA, paired with their corresponding images obtained using the latent decoder, and other symbols are as defined in~\eqref{equ_pt_tx}.
A key distinction between~\eqref{equ_t_ka} and~\eqref{equ_pt_tx} is that the optimization in T-KA is performed over the latents generated by the edge GAI after G-KA, i.e., $\cS_{\rEG}$, as well as under the actual edge-to-user transmission conditions, i.e., $\varPhi_{\reu}$. 
\Copy{R2-0-2}{\rev{Optimizing over the set $\cS_{\rEG}$ in~\eqref{equ_t_ka} couples the T-KA with G-KA, so that they can jointly tackle the original problem~\eqref{equ_ka_prob}.}}

\subsection{\makd for G-KA}\label{s2ec_makd}

Directly solving G-KA problem~\eqref{equ_g_ka} resembles the \emph{DreamBooth} method~\cite{Ruiz23CVPR_Dreambooth}, which fine-tunes the parameters of LDM for GAI personalization.
However, in~\eqref{equ_g_ka}, only the compact LoRA matrices can be trained. 
Due to the severely limited number of parameters of LoRA matrices, distilling sufficient knowledge into them is challenging.

\Copy{R2-2c-1}{\rev{Several recent works enhance LoRA from different perspectives, including improving its training strategy (e.g., Meta-LoRA~\cite{Li25COLING_META_LoRA} and SaRA~\cite{Hu25ICLR_SaRA}) and dynamically expanding its capacity (e.g., SeLoRA~\cite{Mao24Arxiv_SeLoRA}).
Nevertheless, these approaches tend to introduce more complicated training mechanisms or lead to larger LoRA parameter sizes.
In contrast, our proposed method, \makd, aims to minimize the misalignment in semantic interpretation of the target subject between the edge and cloud GAI models, without modifying the training mechanism for LoRA matrices; consequently, our method is orthogonal to existing LoRA enhancement techniques.
}}

More specifically, \makd is based on the insight that the discrepancies between images generated by the cloud-GAI and the edge-GAI arise from two sources: 
(i) the misaligned knowledge for generating details (e.g., textures, patterns, and accessories); and 
(ii) more importantly, model-dependent prompt interpretation, i.e., the same subject may require different prompts across GAI models.
Accordingly, to mitigate the difficulty of distilling knowledge into compact LoRA matrices, \makd prepends a \emph{metaword} $\bm\mu$, i.e., a synthetic token with trainable embedding, to aid the prompt text (e.g., ``a Shiba plush toy'' becomes ``$\bm\mu$ a Shiba plush toy'').
The embedding of $\bm\mu$ is trained to minimize the noise prediction MSE over $\cS_{\rCG}$, which can be expressed as
\ifSingleColumn
\begin{align}\label{equ_metaword_opt}
\min_{\bm\mu}~\sum_{\bmz\in\cS_{\rCG}}\expect\limits_{t,\bm\epsilon}\Big[\|\bm\epsilon - \bm\varepsilon(\bmz_t, \bm\mu\oplus\bmv, t; \varTheta_{\rG})\|^2\Big],
\end{align}
\else
\begin{align}\label{equ_metaword_opt}
\min_{\bm\mu}~\sum_{\bmz\in\cS_{\rCG}}\expect\limits_{t,\bm\epsilon}\Big[\|\bm\epsilon - \bm\varepsilon(\bmz_t, \bm\mu\oplus\bmv, t; \varTheta_{\rG})\|^2\Big],
\end{align}
\fi
where $\oplus$ denotes the concatenation operator.

\Copy{R2-2c-2}{\rev{The optimization in~\eqref{equ_metaword_opt} is inspired by the \emph{textual-inversion} in~\cite{Gal22NIPS_Image}. 
In contrast to~\cite{Gal22NIPS_Image}, where the subject is represented only by the metaword, we concatenate the metaword to the prompt so that the rich semantic content in the prompt is preserved. 
Moreover, we combine it with LoRA so as to enable both prompt-level and parameter-level adaptation.
To the best of our knowledge, such a joint design has not been explored in prior work, particularly in the context of GSC.}}
Conceptually, training the metaword is similar to reprogramming the cloud-generated images into a semantic prefix.
Such a technique is also shown effective in facilitating large language models' understanding of high-dimensional data~\cite{Sheng25WCL_Beam}.

Solving~\eqref{equ_metaword_opt} enables the metaword to minimize the discrepancy between edge- and cloud-generated outputs caused by differences in their semantic interpretation of the prompt text.
As a result, the LoRA matrices only need to compensate for the residual generation knowledge discrepancy.
In this case, the training of the LoRA matrices can be expressed as
\ifSingleColumn
\begin{align}\label{equ_lora_opt}
\min_{\Delta\varTheta_{\rG}}~\sum_{\bmz\in\cS_{\rCG}}\expect\limits_{t,\bm\epsilon}\Big[\|\bm\epsilon - \bm\varepsilon(\bmz_t, \bm\mu^*\oplus\bmv, t; \varTheta_{\rG}+\Delta\varTheta_{\rG})\|^2\Big],
\end{align}
\else
\begin{align}\label{equ_lora_opt}
\min_{\Delta\varTheta_{\rG}}~\sum_{\bmz\in\cS_{\rCG}}\expect\limits_{t,\bm\epsilon}\Big[\|\bm\epsilon - \bm\varepsilon(\bmz_t, \bm\mu^*\oplus\bmv, t; \varTheta_{\rG}+\Delta\varTheta_{\rG})\|^2\Big],
\end{align}
\fi
where $\bm\mu^*$ represents the optimized metaword by~\eqref{equ_metaword_opt}.

\Copy{R2-2c-3}{\rev{Notably, \makd is inherently scalable and readily applicable to multi-subject G-KA.
Specifically, we can optimize a separate metaword for each subject while training a single shared set of LoRA matrices across all subjects. 
This eliminates the need to train and transmit subject-specific LoRA matrices, thereby reducing both computational and communication overhead.}}

\subsection{\rev{\cala for T-KA}}\label{ssec_alg_tka}
\Copy{R2-2a-1}
{\begin{revenv}
The main challenge of the T-KA problem in~\eqref{equ_t_ka} is that simultaneously adapting the JSCC codec to all transmission conditions may lead to suboptimal performance due to interference between discrepant knowledge components.
For example, the transmission knowledge required to align with high-SNR conditions often differs from that for low-SNR conditions.
While such discrepant knowledge components can coexist when the model capacity is sufficient, they tend to interfere with each other in compact LoRA matrices.

A possible approach is to partition the diverse transmission conditions into multiple groups and train a separate set of LoRA matrices for T-KA of each group.
However, this approach suffers from two drawbacks.
First, when the number of potential transmission conditions is large, it becomes nontrivial to group them properly.
Second, under a fixed size constraint of $\Delta\varTheta_{\rJ}$, using multiple sets of LoRA matrices inevitably reduces the capacity of each set, which may in turn degrade the effectiveness of T-KA.

To address the first limitation, we can borrow the gating mechanism from mixture-of-experts (MoE)~\cite{moe_ref} and employ a lightweight neural model to learn the grouping of transmission conditions.
Furthermore, to prevent the capacity reduction caused by the conventional hard on-or-off gating, we propose a fine-grained soft gating mechanism tailored to LoRA matrices as follows.

Each rank-$R$ LoRA matrix in $\Delta\varTheta_{\rJ}$ is expressed as the sum of $R$ rank-$1$ matrices, each carrying a distinct transmission knowledge component.
Specifically, for each LoRA matrix $\bmM = \bmB\bmA$, we have $\bmM = \sum_{i=1}^{R} b_i a_i$, where $b_i$ and $a_i$ are the $i$-th column vector of $\bmB$ and the $i$-th row vector of $\bmA$, respectively.
We introduce a lightweight soft gating function $\bmg:\bbR^3 \rightarrow \bbR^R$, which maps the three-dimensional transmission condition $\bm\phi$ to the weighting coefficients of the $R$ rank-$1$ matrices. 
Then, a condition-aware LoRA matrix can be obtained by the weighted summation, i.e., $\bmM' = \sum_{i=1}^{R} g_i(\bm\phi)\, b_i a_i$.
Since no rank-$1$ component is explicitly excluded from the summation, the effective capacity is preserved.

\input{sections/algorithm/1_alg}

We denote the set of channel-adaptive LoRA matrices by $\bmG(\Delta\varTheta_{\rJ},\bm\phi)$ and the parameters of all soft gating functions by $\cG$. The T-KA problem in~\eqref{equ_t_ka} is converted to
\beq\label{equ_pt_tx_cala}
\min_{\Delta\varTheta_{\rJ},\cG}~\underset{\bm\phi \sim \varPhi_{\reu}}{\mathbb{E}}\sum_{(\bmx,\bmz)\in \cS_{\rEG}} L(\bmx,\bmz; \varTheta_{\rJ}+\bmG(\Delta\varTheta_{\rJ},\bm\phi)|\bm\phi).
\eeq

Our method, referred to as \cala and given in \eqref{equ_pt_tx_cala}, enables condition adaptiveness by jointly training the parameters of the LoRA matrices and the soft gating functions.
In particular, it mitigates the mutual interference among the knowledge required to align with diverse transmission conditions. 
As the lightweight gating functions introduce only a negligible number of additional parameters, \cala improves the efficiency of T-KA with almost no increase in transmission overhead or computational complexity, as will be shown in Sec.~\ref{s3ec_chnl_adapt}.
\end{revenv}}

Overall, \name can be summarized as Algorithm~\ref{alg_overall}.

%% file: sections/algorithm/1_alg.tex
\begin{figure}[!t]
\begin{algorithm}[H]
\small
\caption{\name, the distillation-enabled KA algorithm for GSC systems.}
\label{alg_overall}
\begin{algorithmic} [1]
\Require Prompt text of subject ($\bmv$), distribution of edge-to-user transmission condition ($\varPhi_{\reu}$), pretrained latent generator and JSCC codec ($\varTheta_{\rG}$, $\varTheta_{\rJ}$), latent encoder and decoder of the edge-GAI ($\bmf_{\rLE}$ and $\bmf_{\rLD}$).
\Ensure Metaword ($\bm\mu^*$), LoRA matrices for latent generator ($\Delta\varTheta_{\rG}^*$), LoRA matrices for JSCC codec ($\Delta\varTheta_{\rJ}^*$), parameters of soft gating functions ($\cG^*$).
\State Use the cloud-GAI to generate image samples for $\bmv$ and encode them with $\bmf_{\rLE}(\cdot)$ to obtain $\cS_{\rCG}$.
\State Prepend $\bm\mu$ to $\bmv$, optimize $\bm\mu$ by~\eqref{equ_metaword_opt}, and then optimize $\Delta\varTheta_{\rG}$ by~\eqref{equ_lora_opt}.~\emph{\#\makd for G-KA}
\State With $\bm\mu^*$, $\Delta\varTheta_{\rG}^*$, and $\bmf_{\rLD}(\cdot)$, generate latent-image pairs for $\cS_{\rEG}$.
\State Establish soft-gating functions for $\Delta\varTheta_{\rJ}$ and optimize them with $\Delta\varTheta_{\rJ}$ by~\eqref{equ_pt_tx_cala}.~\emph{\#\cala for T-KA}
\end{algorithmic}
\end{algorithm}
\vspace{-2ex}
\end{figure}

%% file: sections/5_2_implement.tex
\section{Neural Model Implementation}\label{sec_implement}

\ifSingleColumn
\begin{figure}[t]
    \centering
    \includegraphics[width=1\linewidth]{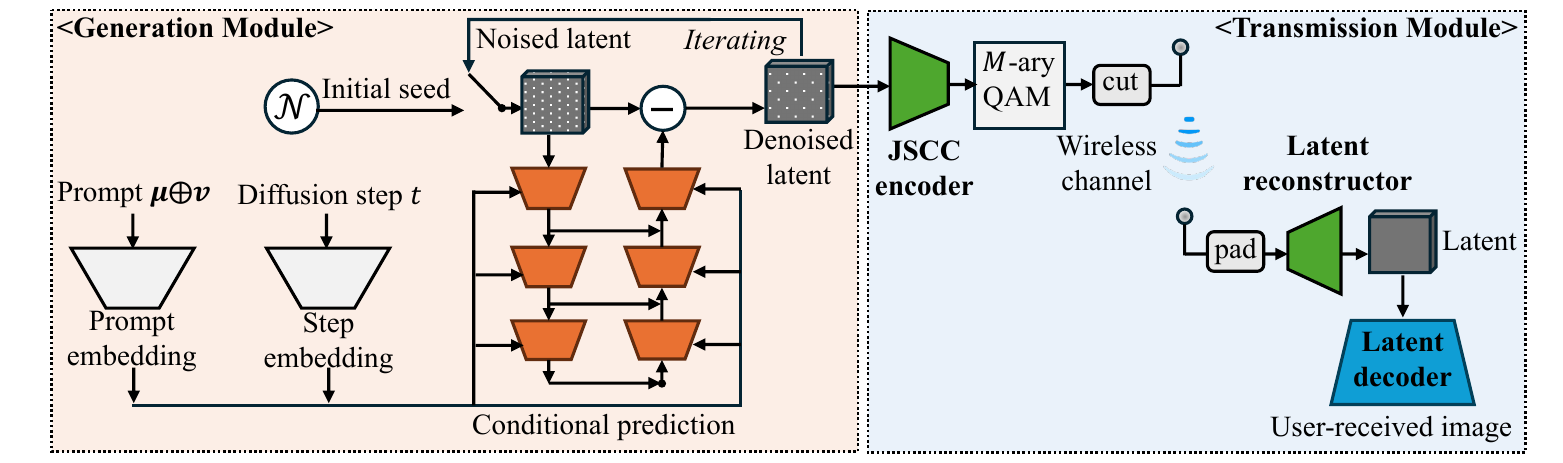}
    \caption{\rev{Designed neural model of the GSC system.}}
    \label{fig_lnm_arch}
\end{figure}
\else
\begin{figure*}[t]
    \centering
    \includegraphics[width=0.85\linewidth]{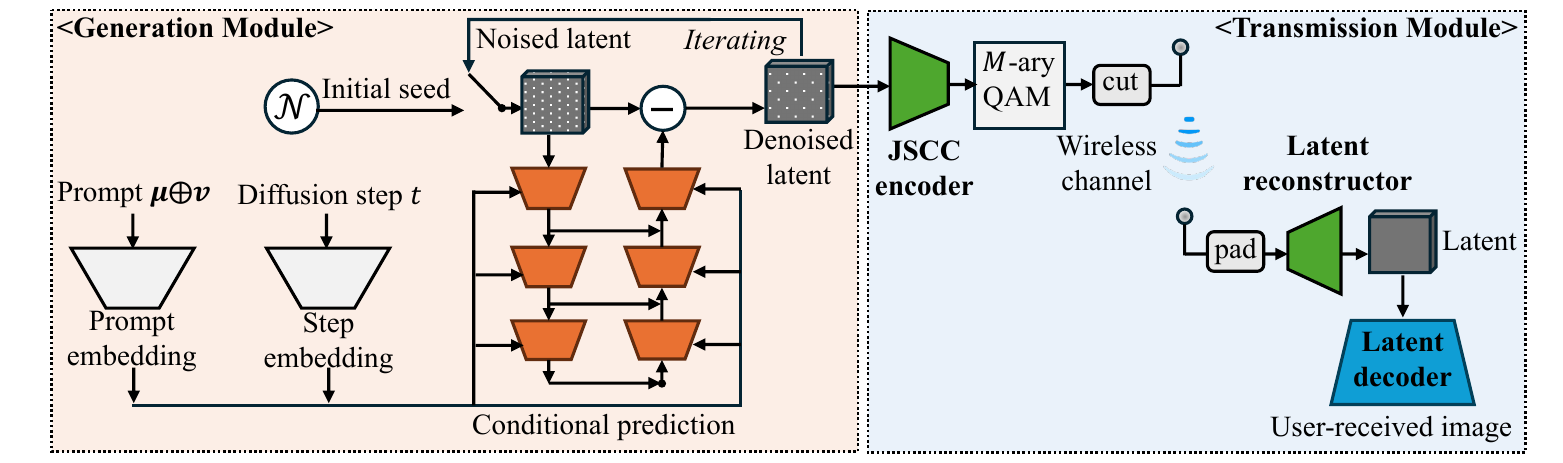}
    \caption{\rev{Designed neural model of the GSC system.}}
    \label{fig_lnm_arch}
\end{figure*}
\fi

To facilitate the evaluation of \name, we design a neural model implementation of the GSC system.
In the following, we describe the implementation of the neural model for the generation and transmission modules.
An overview of the designed neural model is shown in Fig.~\ref{fig_lnm_arch}.

\subsection{Generation Module}\label{s3ec_latent_gen_imp}
In the generation module, the latent space is established by the latent encoder-decoder pair in~\cite{Rombach22CVPR_High}.
The latent encoder takes as input an RGB image $\bmx \in \mathbb{R}^{512 \times 512 \times 3}$ with 8-bit channels, and encodes it into a latent representation $\bmz \in \mathbb{R}^{64 \times 64 \times 4}$ through three downsampling processes.
Conversely, the latent decoder decodes the latent $\bmz$ back to an RGB image of size $512\times 512$.
Both the latent encoder and decoder adopt a hierarchical architecture with convolutional layers to progressively encode/decode local patterns, which incorporate multi-head attention blocks~\cite{Vaswani17NIPS_Attention} to capture global context and residual connections~\cite{He2016_CVPR} to facilitate gradient back-propagation.

As for the noise predictor, it consists of a prompt embedding module, a step embedding module, and a conditional prediction module.

\textbf{Prompt embedding module}: This module first splits the prompt text into words based on spaces and punctuation marks and then tokenizes the words into a sequence of integer indices. 
Subsequently, each token index is mapped to a $1024$-dimensional real-valued vector via a pre-defined embedding matrix, which yields an embedding sequence of the prompt.

\textbf{Step embedding module}: 
This module first embeds each diffusion step index $t$ of the forward/backward diffusion process into a $128$-dimensional vector through sinusoidal position encoding.
The vector is then processed by a multi-layer perceptron with two linear layers, producing a $1280$-dimensional output.

\textbf{Conditional prediction module}: 
This module is based on the U-Net architecture~\cite{unet}, comprising an encoder and a decoder with skip connections that retain multi-scale spatial details.
Taking a $t$-th step noised latent $\bmz_t\in\mathbb R^{64\times 64\times4}$ as input, the encoder progressively extracts three layers of feature maps with decreasing spatial resolutions ($64 \rightarrow 32 \rightarrow 16 \rightarrow 8$) and increasing channel dimensions ($4 \rightarrow 320 \rightarrow 640 \rightarrow 1280$). 
The decoder mirrors the encoder and progressively upsamples the final feature map, ultimately predicting the noise added to $\bmz_t$.
Both the encoding and decoding are conditioned on the step and prompt embeddings.
Specifically, the step embedding is first projected to match the channel dimension and then added to all spatial locations of each feature map.
The prompt embedding is incorporated by flattening each feature map spatially into a sequence of vectors, followed by applying a cross-attention mechanism between this sequence and the embedding sequence of the prompt.

\subsection{Transmission Module}\label{s3ec_jscc_codec}
The JSCC encoder and the the JSCC decoder's latent reconstructor adopt the same hierarchical architecture as the latent encoder and decoder in~\cite{Rombach22CVPR_High} except that only one downsampling layer is used.
In addition, to achieve variable rates, we leverage a truncating-and-padding mechanism as follows.
First, the JSCC encoder maps a latent $\bmz$ to an intermediate tensor in $\mathbb{R}^{64\times 64\times 12}$. 
The target rate is then achieved by truncating along the third (channel) dimension. 
Specifically, given a rate $\tau$, we truncate the last dimension to $128\tau$.
The first and last $64\tau$ channels are treated as the real and imaginary parts, respectively, so that the total number of transmitted symbols satisfies $K = 512\times 512 \,\tau$. 
Before feeding the received symbols into the latent reconstructor, we concatenate the real and imaginary parts and zero-pad along the last dimension back to $\mathbb{R}^{64\times 64\times 12}$ for unified processing. 
With this mechanism, $\tau$ can take values in $\{1/64, 2/64, \ldots, 6/64\}$. 
Finally, the output of the latent reconstructor is passed to the latent decoder.
For implementation efficiency, the latent decoder reuses the one in the generation module.

%% file: sections/6_simulation.tex
\section{Performance Evaluation}\label{sec_eval}

In this section, we first elaborate on the evaluation setup and then provide the evaluation results of the proposed \name algorithm for the GSC system.

\subsection{Evaluation Setup}
In what follows, we describe the evaluation setup of \name in terms of four aspects.
\subsubsection[GAI]{Pre-trained GAI Models}\label{s3ec_pretrain_gai}
We assume the cloud server and the edge server both have LDM-based text-to-image GAI models.
The edge-GAI follows the architecture designed in Sec.~\ref{sec_implement}, adopting the pre-trained parameters of Stable Diffusion 2.1~\cite{stablediffusion21} (SD2.1 for short).
In comparison, the cloud server employs a more powerful GAI model, namely the pre-trained PixArt-$\Sigma$-XL model~\cite{Chen24ECCV_Pixart} (PAXL for short).
The PAXL and SD2.1 use the same latent codec, which has $83.6$ million parameters, occupying $159.6$~MB of storage.
As for the latent generator, PAXL has a significantly larger number of parameters compared to SD2.1 ($5.37$ billion vs. $1.21$ billion), resulting in higher memory consumption ($10.2$~\!GB vs. $2.33$~\!GB) and computational cost. 
As a result, PAXL has richer embedded knowledge for target subjects.
The total number of steps is $1{,}000$, and the number of diffusion steps in generation is $50$.

\Copy{R2-3_1}{
    \rev{Inherently, this configuration coincides with the GAI-as-a-service paradigm in 6G~\cite{Zhou24WCL_Generative}. The knowledge misalignment between the cloud- and edge-GAIs stems not only from their different model sizes, but also from discrepancies in training data distributions, training strategies, and model architectures, as commonly seen in edge-AI settings~\cite{Letaief21JSAC_Edge}.}
}


\ifSingleColumn
\else
\begin{figure*}[t] 
    \centering
    \includegraphics[width=0.9\linewidth]{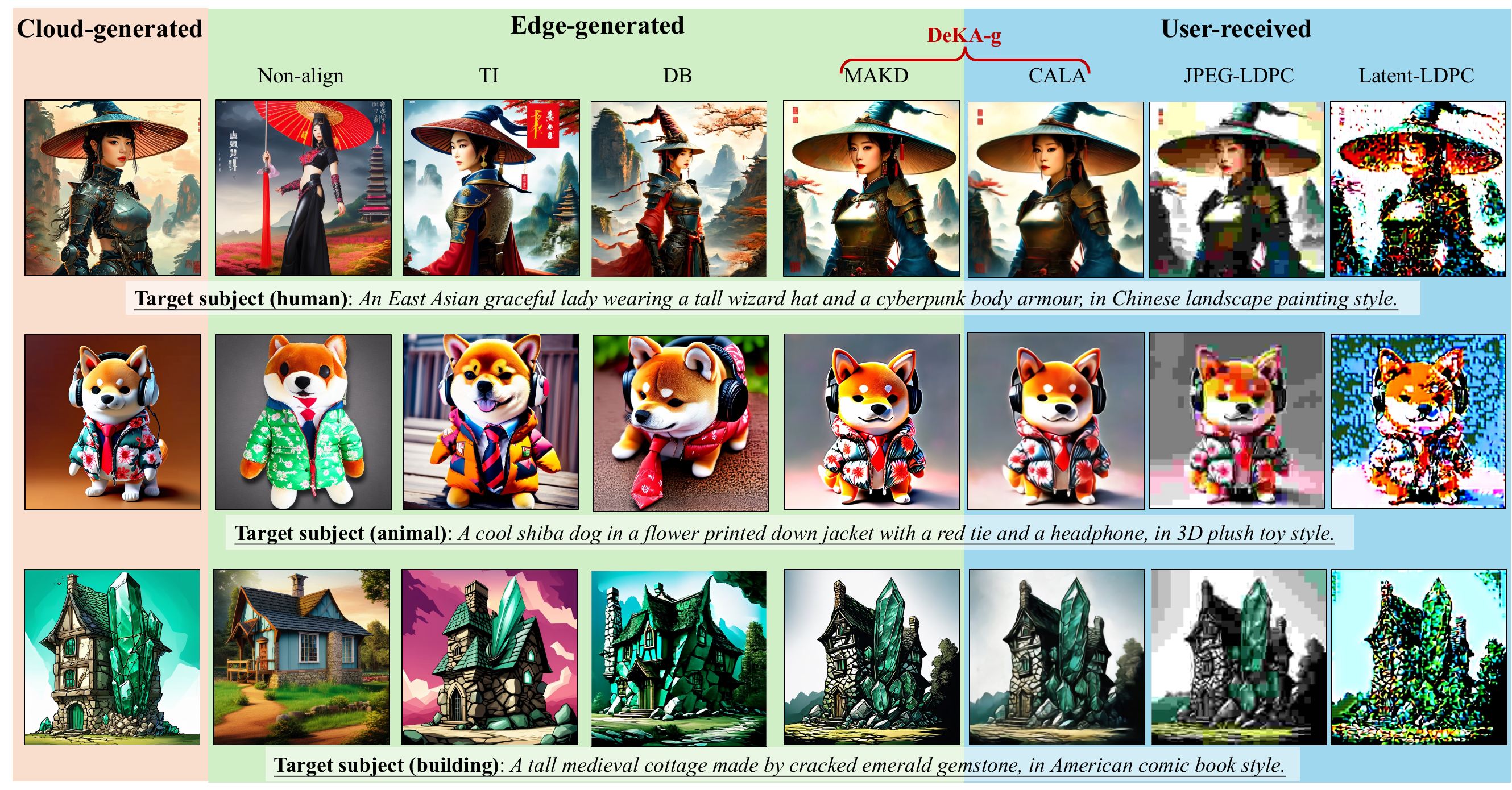}
    \caption{\rev{Overall comparison between \name and benchmark algorithms.}}
    \label{fig1_exp}
\end{figure*}
\fi

\subsubsection{Wireless Channel Conditions}\label{s3ec_chnl_cond}
Based on the typical setup of the urban macro cell~(UMa) scenarios in 3GPP standard~\cite{3GPP_TR_38_901}, we assume a cellular network operating in the $3.5$\!~GHz band with $J=120$ sub-channels allocated to the user.
In the quantization, $M=64$ discrete constellation points are adopted, corresponding to the $64$-ary quadrature amplitude modulation~(QAM).
To focus on the evaluation of KA, we assume perfect channel estimation.
In T-KA, the set of rates is $\tau \in \{i/64\mid i=2,3,...,6\}$, ensuring an integer number of cut-off channels.
Additionally, the set of SNRs is $\gamma\in\{0, 5, \dots, 20\}$\!~dB, and the set of delay spread values is $\omega\in\{30, 300, 3000\}$\!~ns.

\subsubsection{Target Subjects and Sample Sets}
To facilitate an insightful evaluation, we focus on three representative subject categories, including humans, animals, and buildings.
For every category, we randomly select ten target subjects and use their detailed descriptions as the prompt texts.
In G-KA, the cloud-GAI generates $50$ images for each subject's prompt text given different initial seeds. 
Among them, $N_{\rCG}=40$ are used as the sample set $\cS_{\rCG}$, and the remaining $10$ are used to test the consistency scores.
To obtain the latent sample set $\cS_{\rEG}$ for T-KA, we use the edge-GAI model to generate $N_{\rEG}=100$ latents given different initial seeds.
Additionally, a separate set of $30$ latents is generated to test the consistency scores as well as the transmission performance.

\subsubsection{Training Hyperparameters}

In~\eqref{equ_metaword_opt}, the $1024$-dimensional embedding of the meta-word is initialized using a Gaussian distribution with zero mean and variance $0.02$, and then optimized for $4{,}000$ epochs with a learning rate of $5\!\times\!10^{-4}$.
In~\eqref{equ_lora_opt}, zero-initialized LoRA matrices with a uniform rank of $R=8$ are created for all the parameter matrices of the U-Net module.
These LoRA matrices comprise approximately $2.7$ million parameters, resulting in a memory and transmission overhead of only $5.2$\!~MB.
We train these LoRA matrices for $4{,}000$ epochs with a learning rate of $10^{-4}$.

As for the JSCC codec, it comprises approximately $50.5$ million parameters.
In default, the JSCC codec is pre-trained under $10$~\!dB SNR and $300$~\!ns delay spread, using latents extracted from $1{,}000$ images sampled from the DiffusionDB~\cite{wang2023diffusiondb} dataset.
Zero-initialized LoRA matrices with a uniform rank of $R=16$ are created for the JSCC encoder, latent reconstructor, and the first six layers of the latent decoder.
As a result, the total numbers of parameters in the LoRA matrices and in the soft gating functions are $0.6$ million and $2.4$ thousands, respectively, resulting in a memory and transmission overhead of only $1.2$\!~MB.
In~\eqref{equ_pt_tx_cala}, the optimization variables are trained for $1{,}000$ epochs with a learning rate of $10^{-4}$. 

\subsection{Evaluation Results}

We present the evaluation results of the \name algorithm, beginning with an overall comparison, followed by detailed performance analyses of the \makd and \cala methods.

\subsubsection{Overall Comparison}\label{s3ec_overall_eval}
\ifSingleColumn
\begin{figure}[t]
    \Copy{R3-2-fig2}{
    \centering
    \includegraphics[width=1\linewidth]{./figures/eval_res/fig_1_clip_5_0124.pdf}
    \caption{\rev{Overall comparison between \name and benchmark algorithms.}}
    \label{fig1_exp}}
\end{figure}
\else
\fi

In Fig.~\ref{fig1_exp}, we show the overall performance of \name, focusing on three representative subjects in the human, animal, and building categories (additional subjects are reported in the Supplementary Material).
The first column shows the cloud-generated images given the prompt text of each target subject.
Without KA, the same prompt texts result in edge-generated images shown in the \emph{non-align} case, which are substantially different from the cloud-generated ones, visualizing the knowledge misalignment.

For G-KA, we compare \name with the Textual-Inversion~(TI) in~\cite{Gal22NIPS_Image} and the DreamBooth~(DB) in~\cite{Ruiz23CVPR_Dreambooth}, which are the basis of \name and thus serve as important benchmarks. 
TI addresses G-KA by optimizing only the metaword, while DB optimizes the LoRA matrices under the initial random embedding of the metaword.
From Fig.~\ref{fig1_exp}, TI and DB align the edge-generated images with the cloud-generated ones, with TI performing better in layout consistency and DB performing better in detail consistency.
Our \makd method in \name combines the strength of TI and DB and outperforms them in both layout and detail consistency.

\rev{For the wireless transmission from the edge to the user, we compare the \cala method in \name with two communication-centric baselines:
i) the \emph{JPEG-LDPC} baseline, where the generated images are first compressed using Joint Photographic Experts Group~(JPEG) and then transmitted via LDPC coding; and
ii) the \emph{Latent-LDPC} baseline, where the latent tensors generated by the edge-GAI are directly transmitted using standard Low-Density Parity-Check~(LDPC) channel coding.
The transmission condition is characterized by rate $\tau=1/32$, a SNR of $\gamma=10$~dB, and a delay spread value of $\omega=3000$~ns.
We evaluate all methods under the same number of transmitted symbols-per-pixel, which directly reflects the consumption of wireless spectral resources.
In particular, for the baselines, we sweep over different combinations of QAM modulation orders and LDPC coding rates to determine the maximum achievable bit budget under a block error rate constraint of $1\%$. 
This procedure emulates the adaptive modulation and coding scheme used in practical communication systems.
The Latent- and JPEG-LDPC baselines satisfy the bit budget by adjusting the quantization order and the JPEG compression quality, respectively.}

\rev{From Fig.~\ref{fig1_exp}, the JPEG-LDPC baseline compresses the images by grouping pixels into uniform blocks, which leads to severe aliasing artifacts.
The Latent-LDPC baseline exhibits even more pronounced distortion, because under the low rate condition, quantization leads to severe precision loss.
In contrast, with the \cala method, \name achieves high-fidelity under the same transmission condition.}

\subsubsection{G-KA Performance}\label{s3ec_eval_g_ka}
\ifSingleColumn
    \renewcommand{\figwidth}{0.25\textwidth}
\else
    \renewcommand{\figwidth}{0.48\linewidth}
\fi
\begin{figure}[t]
    \centering
    \subfloat[]{\includegraphics[width=\figwidth]{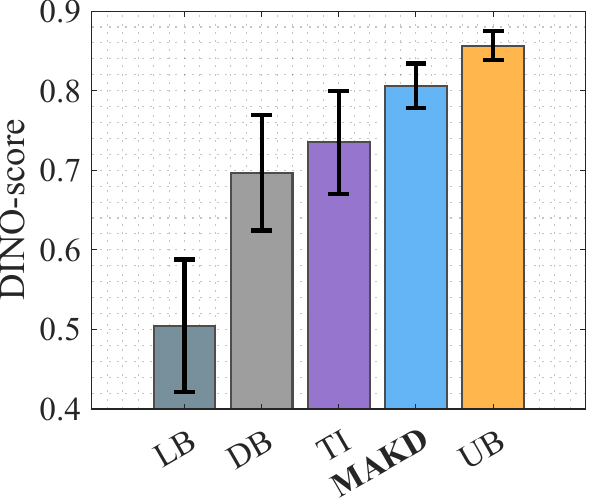}\label{fig:dino_align}}
    \hfill
    \subfloat[]{\includegraphics[width=\figwidth]{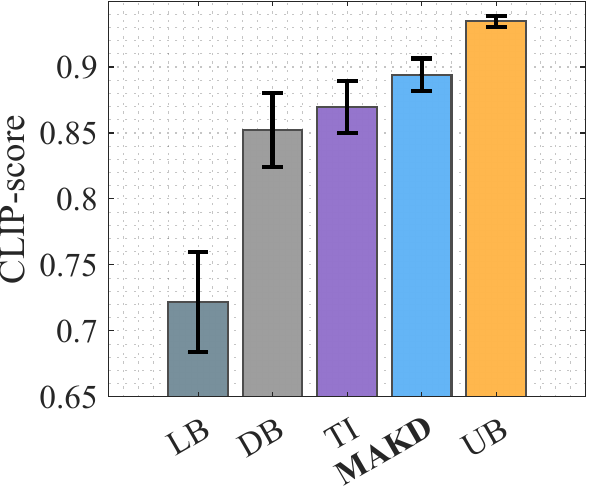}\label{fig:clip_img}}
    \caption{Comparison of different G-KA methods in terms of (a) DINO-score and (b) CLIP-score.}
    \label{fig2_exp}
\end{figure}

In Figs.~\ref{fig:dino_align} and \ref{fig:clip_img}, we compare \makd of \name with TI and DB in terms of the DINO- and CLIP-scores defined in~\eqref{equ_ka_prob}.
We adopt the non-align case as the lower-bound baseline~(LB) and the consistency among cloud-generated images as the upper-bound baseline~(UB).
The average and standard deviation of DINO- and CLIP-scores are evaluated for the test set of each target subject.
Compared with LB, \makd significantly enhances the consistency of user-received images,  with an average improvement of $44$\% in the combined DINO- and CLIP-scores.
Particularly, \makd outperforms the second-best method (TI) by $6.3$\% on average~($9.7\%$ for DINO-score and $2.8\%$ for CLIP-score).
Moreover, compared with TI, \makd reduces the standard deviations of DINO-score and CLIP-score by $56$\% and $38$\%, respectively.
These results highlight the superior effectiveness of \makd in aligning the generation knowledge, which is consistent with the visual results shown in Fig.~\ref{fig1_exp}.
\ifSingleColumn
\begin{figure}[b] 
    \centering
    \includegraphics[width=0.55\linewidth]{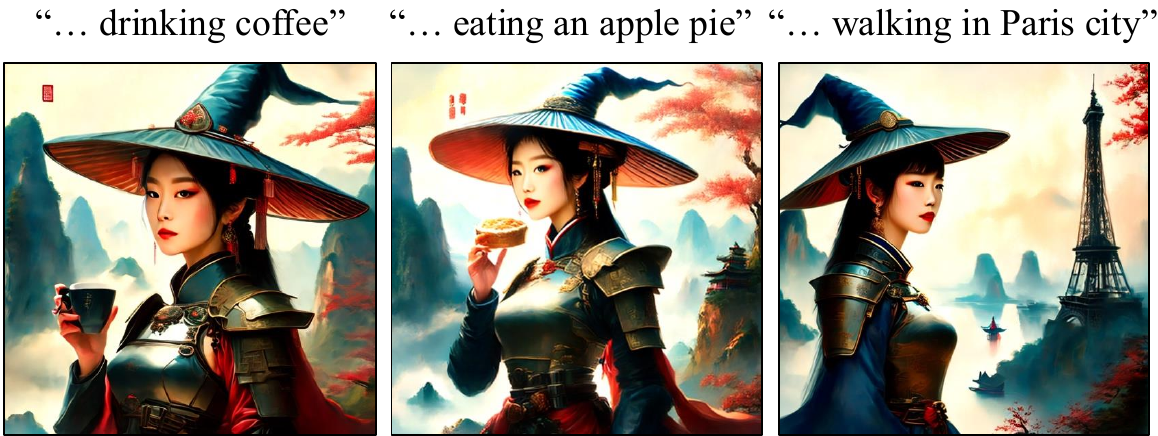}
    \caption{Validation of the generalizability of the G-KA achieved by \name.}
    \label{fig3_exp}
\end{figure}
\else
\begin{figure}[t] 
    \centering
    \includegraphics[width=1\linewidth]{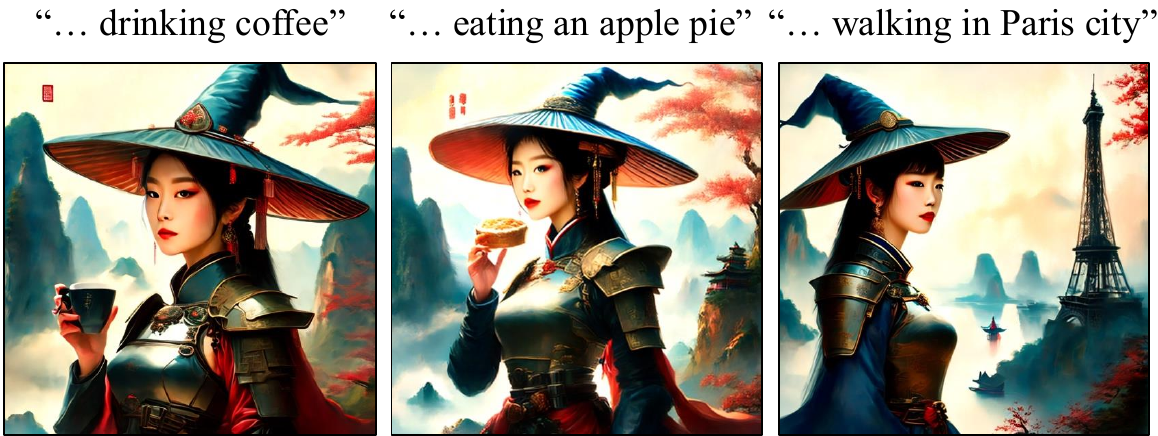}
    \vspace{-1ex}
    \caption{Validation of generalizability of G-KA achieved by \name.}
    \label{fig3_exp}
\end{figure}
\fi

\ifSingleColumn
    \renewcommand{\figwidth}{0.25\textwidth}
    \renewcommand{\manuhspacing}{0.08\linewidth}
\else
    \renewcommand{\figwidth}{0.42\linewidth}
\fi
\begin{figure}[t]
\Copy{R3-2b_1}{
    \centering
    \subfloat[]{\includegraphics[height=\figwidth]{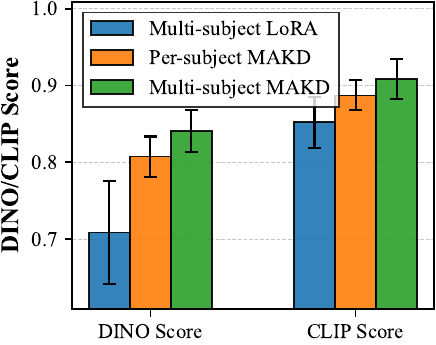}\label{fig_r2_2b_1a}}
    \ifSingleColumn
        \hspace{\manuhspacing} 
    \else
        \hfill
    \fi
    \subfloat[]{\includegraphics[height=\figwidth]{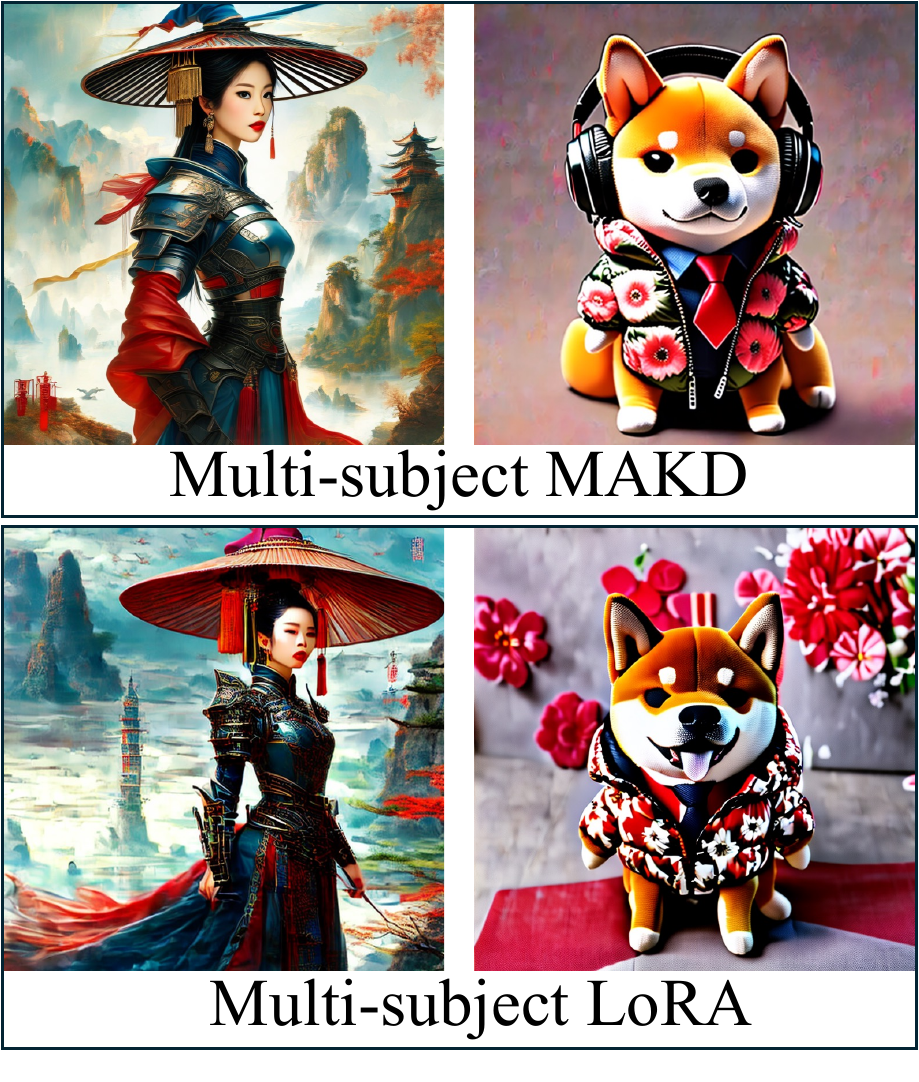}\label{fig_r2_2b_1b}}
    \caption{\rev{Comparison of different methods for two-subject G-KA.
    (a) Quantitative results in terms of DINO score and CLIP score.
    (b) Visual comparison of the generated images.}}
    \label{fig_r2_2b_1}}
\end{figure}

Moreover, we validate the generalizability of G-KA achieved by \makd on extended prompt texts.
As shown in the upper line of Fig.~\ref{fig3_exp}, we extend the metaword-aided prompt text by appending short phrases that describe the subject's interaction with other objects.
Using the edge-GAI model aligned by \makd, we generate latents for these extended prompt texts and decode them into images.
From the images shown in Fig.~\ref{fig3_exp}, the G-KA achieved by \makd demonstrates strong generalizability to extended prompts, successfully generating consistent images in which the target subject interacts appropriately with the described objects.

\Copy{R2-2c-4}{\rev{Furthermore, in Fig.~\ref{fig_r2_2b_1}, we demonstrate the scalability of \makd to multi-subject G-KA.
We evaluate the following three cases:
(i) \emph{Multi-subject LoRA}: a single set of LoRA matrices is trained for G-KA of two target subjects (the human and animal subjects shown in Fig.~\ref{fig1_exp});
(ii) \emph{Per-subject \makd}: the G-KA of each target subject is handled individually by using \makd;
(iii) \emph{Multi-subject \makd}: subject-specific metawords are optimized, after which a shared set of LoRA matrices is trained for both subjects.}

\rev{Fig.~\ref{fig_r2_2b_1a} shows that Multi-subject \makd outperforms the Multi-subject LoRA in terms of both DINO- and CLIP-score, with the resulting images illustrated in Fig.~\ref{fig_r2_2b_1b}.
This indicates that optimized metawords can improve the effectiveness of LoRA for G-KA.
Moreover, comparing Multi-subject \makd with Per-subject \makd shows that with metawords, sharing a single set of LoRA matrices yields better performance than using separate LoRA matrices for each subject.
This is possibly because, once the discrepancy in semantic interpretation is mitigated by metawords, a shared set of LoRA matrices can more effectively capture the common generation knowledge across targets than separate ones.
This result demonstrates the scalability of \makd to multi-subject G-KA.}}

\ifSingleColumn
    \renewcommand{\figwidth}{0.7\textwidth}
\else
    \renewcommand{\figwidth}{1\linewidth}
\fi
\begin{figure}
\Copy{R1_4_fig1}{
\centering
\includegraphics[width=\figwidth]{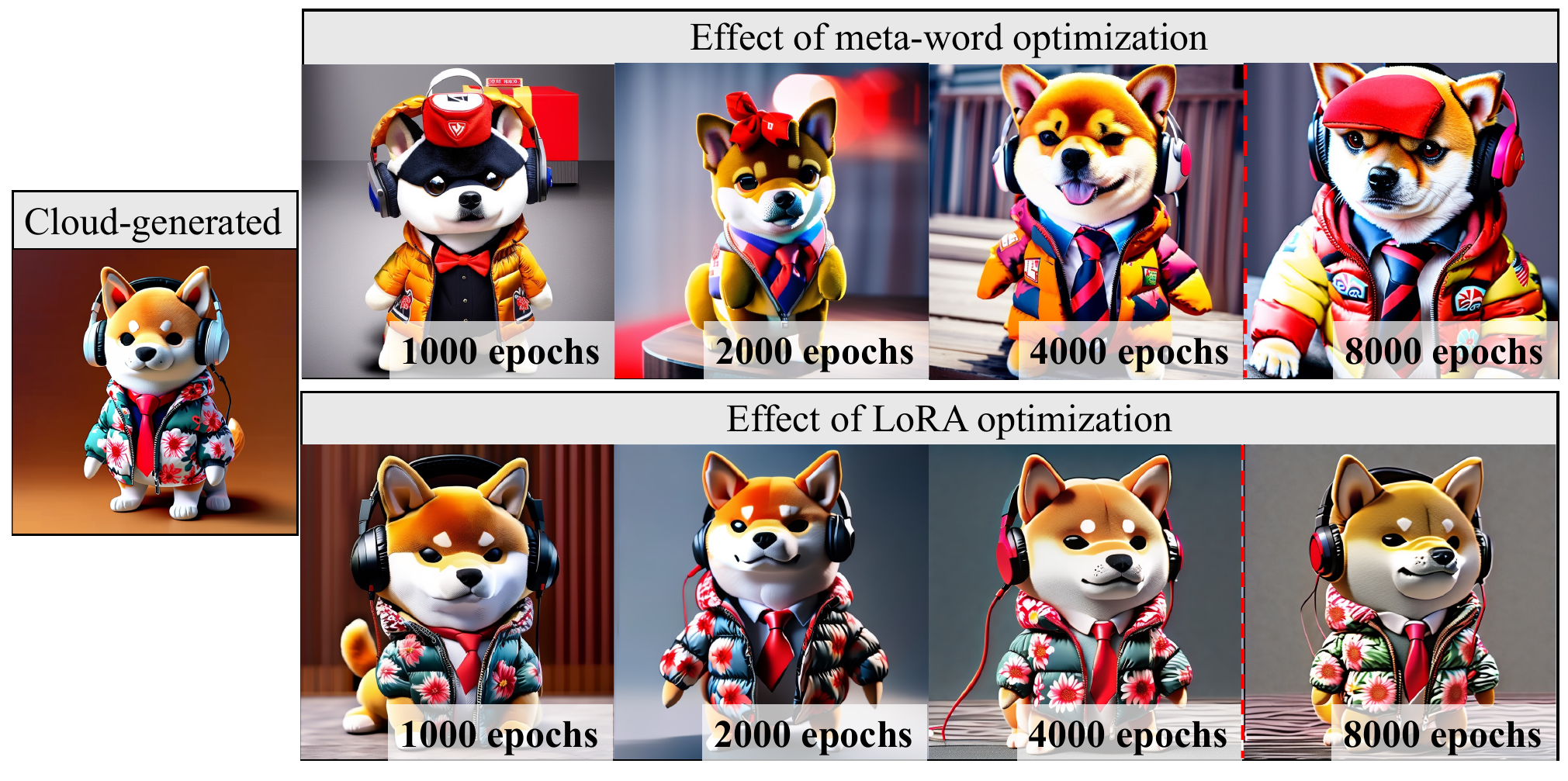}
\vspace{-2ex}
\caption{\rev{Impact of inappropriate training duration in G-KA on the edge-generated images}.}
\label{R1_4_fig1}
}
\end{figure}

\Copy{R1-4-1}{\rev{To illustrate the limitations of the proposed algorithm, we examine G-KA failure cases under inappropriate training durations and severely limited distilled knowledge capacity.
Fig.~\ref{R1_4_fig1} illustrates the edge-generated results obtained under different numbers of training epochs for the meta-word optimization and the LoRA training processes.
Overall, insufficient training leads to inadequate generation knowledge alignment, resulting in generated images that are not consistent with the target cloud-generated ones.
Moreover, excessive training can also lead to failure cases with deteriorated consistency.
This is mainly because these training processes use a limited set of cloud-generated images, and overly-deep training on such a small dataset causes overfitting. 
Consequently, the training results become overly tailored to these training samples and lose their generalizability.
Therefore, selecting an appropriate training depth for the generation knowledge alignment~(G-KA) is important for its success.}}

\ifSingleColumn
    \renewcommand{\figwidth}{0.7\textwidth}
\else
    \renewcommand{\figwidth}{1\linewidth}
\fi
\begin{figure}
\Copy{R1_4_fig2}{
\centering
\includegraphics[width=\figwidth]{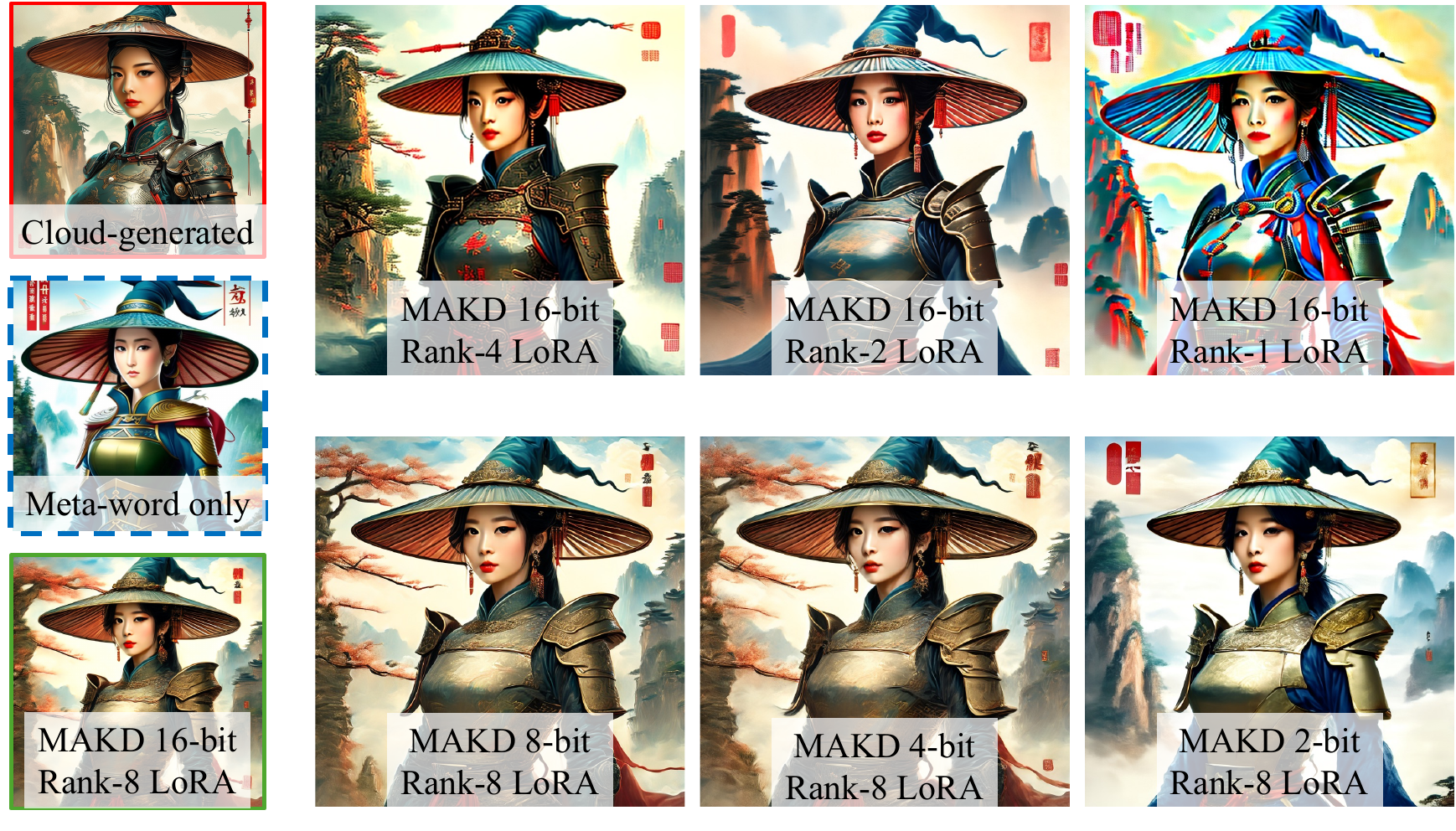}
\vspace{-2ex}
\caption{\rev{Impact of highly-limited LoRA size in G-KA on the edge-generated images}.}
\label{R1_4_fig2}
}
\end{figure}

\Copy{R1-4-2}{\rev{Next, we analyze the failure cases due to severely limited distilled knowledge size.
In the leftmost column of Fig.~\ref{R1_4_fig2}, we show, from top to bottom, the cloud-generated target image, the edge-generated result using only the optimized metaword, and the edge-generated result obtained by \makd using rank-8 LoRA matrices. 
Incorporating LoRA-based distilled knowledge clearly improves the similarity between edge- and cloud-generated images.
However, when the size of the distilled knowledge is further constrained, G-KA degrades or even fails. 
As shown in the first row on the right-hand side of Fig.~\ref{R1_4_fig2}, reducing the rank budget to $4$, $2$, and $1$ progressively results in less aligned images. 
In particular, with rank-1 LoRA, severe distortions are observed, indicating that overly stringent size constraints on distilled knowledge can lead to failures.}

\rev{To address this issue, we can reduce the size of the distilled knowledge via lowering the precision rather than lowering the rank budget.
In particular, rank-$8$ LoRA matrices in 16-bit precision can be quantized to $8$-bit, $4$-bit, or $2$-bit precision, resulting in a sizes close to that of rank-$4$, rank-$2$, or rank-$1$ LoRA matrices, respectively.
The second row on the right-hand side of Fig.~\ref{R1_4_fig2} shows the edge-generated images obtained with the distilled knowledge under different precision levels.
It can be observed that even under a mere 2-bit precision, the proposed method still yields a clear improvement in similarity between cloud- and edge-generated images, demonstrating a more favorable trade-off between distilled knowledge size and G-KA performance than reducing the rank budget.}}

\subsubsection{T-KA Performance}\label{s3ec_chnl_adapt}
\ifSingleColumn
\renewcommand{\figwidth}{0.32\textwidth}
\else
\renewcommand{\figwidth}{0.28\linewidth}
\fi

\ifSingleColumn
\begin{figure}[btp]
\else
\begin{figure*}[btp]
\fi
\Copy{R2_2a_fig1}{
    \centering
    \subfloat[]{\includegraphics[width=\figwidth]{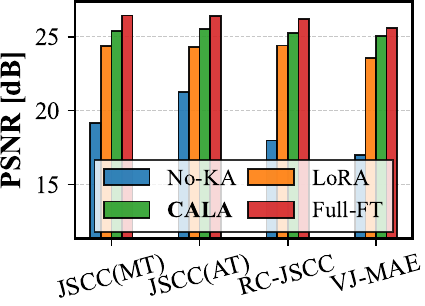}\label{R2_2a_fig1_psnr_loss}}
    \subfloat[]{\includegraphics[width=\figwidth]{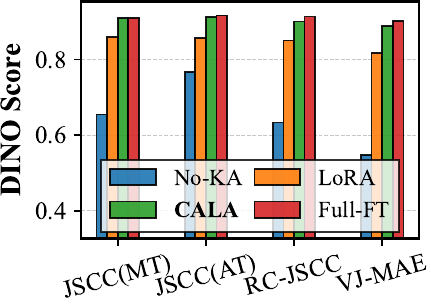}\label{fig6:R2_2a_fig1_dino_loss}}
    \subfloat[]{\includegraphics[width=\figwidth]{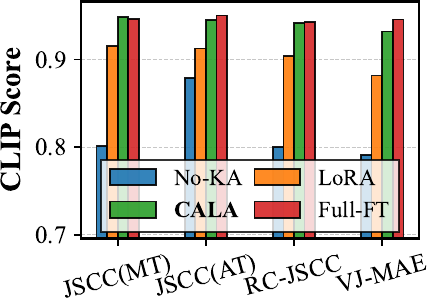}\label{R2_2a_fig1_clip_loss}}
    \caption{\rev{Comparison of T-KA methods in different cases of pretrained JSCC codecs, measured by (a) PSNR, (b) DINO-score, and (c) CLIP-score.}}}
    \label{R2_2a_fig1}
\ifSingleColumn
\end{figure}
\else
\end{figure*}
\fi
\Copy{R2-2a-2}{\rev{We first evaluate the effectiveness of the proposed \cala method for T-KA, applied to the JSCC codec in Sec.~\ref{s3ec_jscc_codec} as well as other benchmark JSCC codecs.
Specifically, we evaluate four cases of JSCC codecs, including {JSCC(MT)}, {JSCC(AT)}, {RC-JSCC}, and {VJ-MAE}.
Both the {JSCC(MT)} and {JSCC(AT)} cases adopt the JSCC codec designed in Sec.~\ref{s3ec_jscc_codec}.
{JSCC(MT)} is pretrained under \emph{misaligned transmission conditions} with the actual transmission condition $\varPhi_{\reu}$. 
In comparison, {JSCC(AT)} is pretrained under \emph{aligned transmission conditions} with $\varPhi_{\reu}$; therefore only the latent data distribution is misaligned.}

\rev{Cases {RC-JSCC} and {VJ-MAE} adopt the variable-rate JSCC codec designs in \cite{Yang22ICASSP_JSCC} and \cite{Wang25COMM_Joint}, respectively.
These two works are selected as benchmarks because \cite{Yang22ICASSP_JSCC} is a widely cited variable-rate scheme, while \cite{Wang25COMM_Joint} represents one of the latest approaches.
For a fair comparison, in both the {RC-JSCC} and {VJ-MAE} cases, we allow the use of the same latent decoder as that in {JSCC(MT)/(AT)} while following \cite{Yang22ICASSP_JSCC} and \cite{Wang25COMM_Joint} to transmit and reconstruct latents over wireless channels.  
Moreover, in the pretraining of both cases, the same SNR and delay spread distributions as in $\varPhi_{\reu}$ are adopted, so that only the rate distribution is misaligned.
Regarding the rate, {RC-JSCC} follows \cite{Yang22ICASSP_JSCC} by optimizing a rate-distortion objective to adapt the rate to the SNR, whereas {VJ-MAE} adopts the strategy in \cite{Wang25COMM_Joint}, where the transmission rate is determined by the entropy of the latent to transmit.}

\ifSingleColumn
    \renewcommand{\figwidtha}{0.22\linewidth}
    \renewcommand{\figwidthb}{0.22\linewidth}
    \renewcommand{\figwidthc}{0.22\linewidth}
\else
    \renewcommand{\figwidtha}{0.33\linewidth}
    \renewcommand{\figwidthb}{0.32\linewidth}
    \renewcommand{\figwidthc}{0.32\linewidth}
\fi
\begin{figure}[bp]
    \Copy{R3_1_fig3}{
    \centering
    \subfloat[]{\includegraphics[width=\figwidtha]{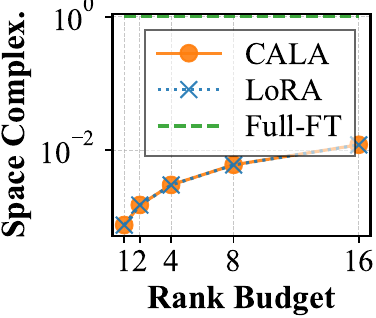}\label{R3_1_fig3_a}}
    \ifSingleColumn
    \hspace{3ex}
    \else
    \hfill
    \fi
    \subfloat[]{\includegraphics[width=\figwidthb]{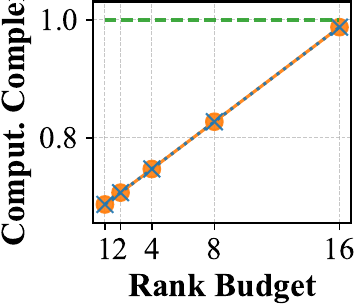}\label{R3_1_fig3_b}}
    \ifSingleColumn
    \hspace{3ex}
    \else
    \hfill
    \fi
    \subfloat[]{\includegraphics[width=\figwidthc]{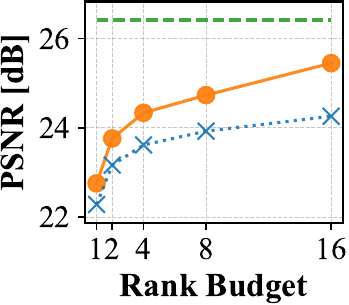}\label{R3_1_fig3_c}}
    \caption{\rev{Comparison of (a) space complexity, (b) computational complexity, and resulting PSNR among different T-KA methods. The legend shown in (a) applies to all subfigures.}}
    \label{R3_1_fig3}}
\end{figure}

\rev{Under the four cases of JSCC codecs, we evaluate the average PSNR, DINO-score, and CLIP-score of four T-KA methods:
i) \emph{No-KA}: the pretrained JSCC codec is directly evaluated under the test setup, i.e., transmitting edge-generated latents after G-KA under the actual transmission conditions;
ii) \emph{LoRA}: a set of LoRA matrices is trained for T-KA across diverse transmission conditions;
iii) \emph{\cala}: a set of condition-aware LoRA matrices are trained for T-KA; and
iv) \emph{Full-FT}: all parameters of the JSCC codec are fine-tuned for T-KA.}

\rev{As shown in Fig.~\ref{R2_2a_fig1}, both {LoRA} and {\cala} substantially improve the performance of all JSCC codecs under the test setup while {\cala} consistently outperforms {LoRA} and approaches the performance of {Full-FT}.
Regarding the DINO and CLIP scores, {\cala} consistently outperforms {LoRA} and {No-KA}, achieving performance on par with {Full-FT}.
In terms of PSNR performance, {\cala} achieves as large as 6.5~\!dB gain over {No-KA}, only 0.8\!~dB lower than {Full-FT}, but with transmitting only about 1\% of the total neural parameters of a JSCC codec.
Compared to LoRA, it achieves a 1.2~\!dB gain in terms of PSNR.}}

\ifSingleColumn
    \renewcommand{\figwidth}{0.32\textwidth}
\else
    \renewcommand{\figwidth}{0.48\linewidth}
\fi
\begin{figure}[bp]
\Copy{R3_1_fig1}{
    \centering
    \subfloat[]{\includegraphics[width=\figwidth]{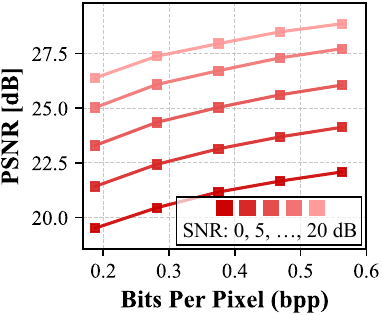}\label{R3_1_fig1_a}}
    \subfloat[]{\includegraphics[width=\figwidth]{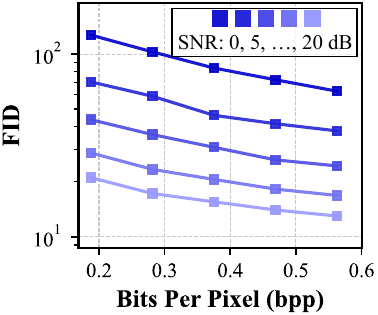}\label{R3_1_fig1_b}}
    \caption{\rev{Bitrate-distortion trade-offs of the JSCC codec after the T-KA using \cala, under different SNRs: (a) PSNR versus bpp; (b) FID versus bpp.}}
    \label{R3_1_fig1}}
\end{figure}

\ifSingleColumn
    \renewcommand{\figwidth}{0.32\textwidth}
\else
    \renewcommand{\figwidth}{0.24\linewidth}
\fi
\ifSingleColumn
\begin{figure}[tp]
\else
\begin{figure*}[t]
\fi
    \Copy{R3_2_fig3}{
    \centering
    \subfloat[]{\includegraphics[width=\figwidth]{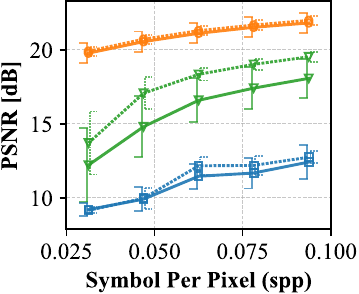}\label{R3_2_fig3_a}}
    \hspace{2ex}
    \subfloat[]{\includegraphics[width=\figwidth]{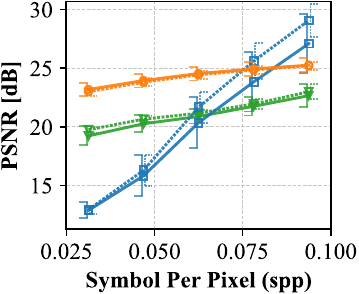}\label{R3_2_fig3_b}}
    \hspace{2ex}
    \subfloat[]{\includegraphics[width=\figwidth]{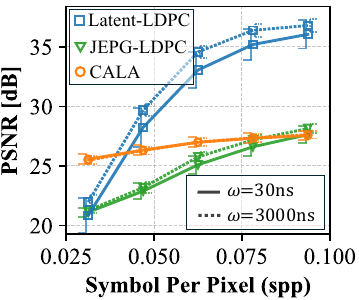}\label{R3_2_fig3_c}}
    \caption{\rev{Comparison of the proposed method (i.e., JSCC after T-KA) with two baselines under different SNR, delay spreads, and symbol-per-pixel settings, measured in terms of PSNR. (a) Low SNR, $\gamma = 0$ dB; (b) Medium SNR, $\gamma = 10$ dB; (c) High SNR, $\gamma = 20$ dB. The legend shown in (c) applies to all subfigures.}}
    \label{R3_2_fig3}}
\ifSingleColumn
\end{figure}
\else
\end{figure*}
\fi

\Copy{R3-1-3}{\rev{In Fig.~\ref{R3_1_fig3_a}, we compare the space complexity, computational complexity, and the resulting PSNR among LoRA, \cala, and Full-FT for T-KA, under different rank budgets for the update parameters.
Notably, Full-FT is unaffected by the rank budget, as it fine-tunes all parameters.
The space complexity is measured by the ratio of the update parameter size to that of the full JSCC codec.
The computational complexity is measured by the ratio of the number of floating-point operations~(FLOPs) required for training to those required for full fine-tuning.
As shown in Fig. \ref{R3_1_fig3_a}, \cala achieves a reduction of more than two orders of magnitude in space complexity compared to Full-FT. Moreover, when the rank budget is smaller than $16$, the training computational complexity is also lower than that of Full-FT. 
Compared with LoRA, \cala achieves higher PSNR with nearly identical space and computational complexity, demonstrating the enhanced T-KA performance due to channel awareness.}}

\Copy{R3-1-1}{\rev{Next, we evaluate the bitrate-distortion trade-offs of the JSCC codec after the T-KA by using the \cala method.
Specifically, we evaluate the relationship between bits per pixel (bpp) and PSNR as well as Fr\'echet Inception Distance~(FID), which are standard quality metrics in image communication systems.
Here, the bpp is computed as $6$ times the number of complex symbols per pixel, since 64-QAM is adopted in Sec.~\ref{s3ec_chnl_cond} and thus each symbol carries $6$ bits.
As shown in Figs.~\ref{R3_1_fig1_a} and~\ref{R3_1_fig1_b}, increasing bpp consistently leads to higher PSNR and lower FID, demonstrating a monotonic improvement in image quality as more spectral resources are used.
Notably, the bpp of the proposed method falls in the $0.2\ssim 0.6$ range, which is much lower than the $24$ bpp of an RGB image, indicating substantial compression.}}

\Copy{R3-1-2}{\rev{As the SNR is the dominant wireless condition affecting the transmission, we analyze the sensitivity of the proposed method to SNR.
In Figs.~\ref{R3_1_fig1_a} and~\ref{R3_1_fig1_b}, we compare the bpp-PSNR and bpp-FID curves under different SNRs. 
Over the SNR range of $0\ssim20$~\!dB, the PSNR degrades by approximately 0.34~\!dB per 1~\!dB decrease in SNR, while the FID increases by about 9.0\% per 1~\!dB decrease in SNR.
Therefore, no abrupt performance drop is observed as the SNR decreases, indicating that the proposed scheme is relatively robust to variations in wireless channel conditions.}}

\Copy{R3-2-1}{\rev{Finally, we compare the proposed method with the two communication-centric baselines methods described in Sec.~\ref{s3ec_overall_eval}, including \emph{Latent-LDPC} and \emph{JPEG-LDPC}.}
\rev{Figs.~\ref{R3_2_fig3_a}, \ref{R3_2_fig3_b}, and \ref{R3_2_fig3_c} show the mean and standard deviation of PSNR for the proposed method (i.e., JSCC after applying T-KA) and the two baselines at SNRs of $0$, $10$, and $20$~dB, respectively.
The left-hand solid-line error bar corresponds to the standard deviation under delay spread $\omega = 30$~ns, whereas the right-hand dotted-line error bar corresponds to that under $\omega = 3000$~ns.
It can be observed that the proposed method outperforms the JPEG-LDPC under most conditions. 
Compared with the Latent-LDPC, the proposed method shows a clear advantage at low-to-medium SNRs, particularly when the spectral resource is highly limited.
In contrast, under high-SNR conditions, or under medium-SNR conditions with abundant spectral resources, the Latent-LDPC achieves better performance.
Regarding the impact of delay spread, a smaller $\omega$ (i.e., stronger channel correlation) generally leads to larger performance variance and a lower mean PSNR. 
Compared with the two baselines, the proposed method is less affected by channel correlation, indicating improved robustness.}}

%% file: sections/7_conclusion.tex
\section{Conclusion}\label{sec_conclu}

In this paper, we have proposed \name, an algorithm to solve the KA problem in GSC systems.
First, we have modeled the GSC system for AIGC image provisioning.
Then, the KA problem is formulated to maximize the consistency between the user-received images and the target AIGC images to be provided by the cloud. 
\rev{To solve the problem, \name comprises the \makd and \cala methods for efficiently aligning generation and transmission knowledge. 
Overall, compared with full-parameter fine-tuning, \name requires transmitting only $0.2$\% of the edge-GAI parameters and about $1$\% of the JSCC-codec parameters.}
Under such compact parameter updates, evaluation results validate that \makd enables a $44$\% increase in the sum of DINO- and CLIP-scores between edge-generated and cloud-generated images, outperforming the second-best benchmark by $6.3$\%. 
\rev{Moreover, \cala enhances PSNR by 6.5~\!dB over the baselines without KA and outperforms conventional LoRA by 1.2~dB.}